\documentclass[review]{elsarticle}
\usepackage[utf8]{inputenc}
\usepackage{amsmath}
\usepackage{array}
\usepackage{graphicx}
\usepackage[hyphens]{url}
\usepackage{multirow}
\usepackage{hyperref}
\usepackage{float}
\usepackage{subfig}
\usepackage{multicol}
\usepackage{mathtools}
\usepackage{schemata}
\usepackage{dsfont}
\usepackage{tabularx}
\usepackage{todonotes}
\usepackage{algorithm} 
\usepackage[inline]{enumitem}
\usepackage{booktabs}
\usepackage{amssymb}
\usepackage{listings}
\usepackage{xcolor}
\usepackage[toc,page]{appendix}

\newcommand{\czm}[1]{{\color{black}#1}} 
\newcommand{\cz}[1]{{\color{black}#1}} 
\newcommand{\cris}[1]{{\color{black}#1}} 

\newcommand{\prev}[1]{{\color{black}#1}}
\newcommand{\eugenio}[1]{{\color{black}#1}} 
\newcommand{\geni}[1]{{\color{black}#1}}
\newcommand{\nuevo}[1]{{\color{black}#1}}

\newcommand{\new}[1]{{\color{black}#1}}
\newcommand{\segrev}[1]{{\color{black}#1}}

\begin{document}
\begin{frontmatter}


\title{\segrev{Sentiment Analysis based Multi-person Multi-criteria Decision Making Methodology using Natural Language Processing and Deep Learning for Smarter Decision Aid. Case study of restaurant choice using TripAdvisor reviews}}
\author[author_ugr]{Cristina Zuheros\corref{cor1}}
\ead{czuheros@ugr.es}
\author[author_decsai]{Eugenio Mart\'{i}nez-C\'{a}mara}
\ead{emcamara@decsai.ugr.es}
\author[author_decsai]{Enrique Herrera-Viedma}
\ead{viedma@decsai.ugr.es}
\author[author_decsai]{Francisco Herrera}
\ead{herrera@decsai.ugr.es}
\address{Andalusian Research Institute in Data Science and Computational Intelligence,\\University of Granada,\\18071 \textit{Granada}, Spain.}
\cortext[cor1]{Corresponding author}

\begin{abstract}
Decision making models are constrained by taking the expert evaluations with pre-defined numerical or linguistic terms. We claim that the use of sentiment analysis will allow decision making models to consider expert evaluations in natural language. Accordingly, we propose the \nuevo{Sentiment Analysis based Multi-person Multi-criteria Decision Making (SA-MpMcDM) methodology \segrev{for smarter decision aid}}, which builds the expert evaluations from their natural language reviews\nuevo{, and even from their numerical ratings if they are available}. \nuevo{The SA-MpMcDM methodology incorporates an end-to-end multi-task deep learning model for aspect based sentiment analysis, named DOC-ABSADeepL model, able to identify the aspect categories mentioned in an expert review, and to distill their opinions and criteria. The individual evaluations are aggregated via \new{the procedure \segrev{named }}
criteria weighting through the attention of the experts.} 
\segrev{We evaluate the methodology in a case study of restaurant choice using TripAdvisor reviews, hence we build, manually annotate, and release the TripR-2020 dataset of restaurant reviews.} We \nuevo{analyze the SA-MpMcDM methodology in different scenarios using and not using natural language and numerical evaluations. The analysis shows that the combination of both sources of information results in a higher quality preference vector.}

\end{abstract}

\end{frontmatter}

\section{Introduction}

Decision making (DM) is an essential cognitive process of human beings. Over time, different models have emerged to help us \nuevo{to} solve DM problems. In particular, \nuevo{multi-person} multi-criteria decision making \nuevo{(MpMcDM)} models \nuevo{consider} the evaluations of multiple experts to solve a decision situation analyzing all possible solution alternatives \nuevo{according to} several criteria \cite{yager1993non}.

Computational DM process, as the human DM one, requires of useful, complete and insightful information for making the most adequate decision according to the input information. The input of DM models \nuevo{is} usually a set of evaluations from the experts. 
\nuevo{They} wish to express their evaluations in natural language, but raw text is not directly processed by DM models. Accordingly, several approaches are followed for asking and elaborating a computational representation of the evaluations
, namely: \begin{enumerate*}[label=(\arabic*)]\item using a numerical representation of the evaluations \cite{roubens1997fuzzy} and \item using a pre-defined set of linguistic terms \cite{herrera2000linguistic}.\end{enumerate*} These approaches for asking evaluations constrain the evaluative expressiveness of the experts, because they have to adapt their evaluation to the numerical or linguistic evaluation \nuevo{alternatives}. We claim that experts in a DM problem have to express their evaluations in natural language, and the DM model has to be able to process and computationally represent them.

\nuevo{Natural language processing (NLP) is the artificial intelligence area that combines linguistic and computational language backgrounds for understanding and generating human language \cite{nlp_handbook,cl_handbook}. NLP is composed of several tasks focused on different aspects of language in order to represent it and extract insightful knowledge from it, namely: machine translation \cite{koehn_2020}, argument mining \cite{lawrence2020}, text summarization \cite{torres2014}, information extraction \cite{sarawagi2008}, among others.

The evaluations of experts in DM are the expression of their private states about a set of target alternatives \cite{Quirk1985}. The language used for projecting those private states is subjective language \cite{wiebe2004}. The NLP task concerned with the treatment of opinions, sentiments and subjectivity in text is sentiment analysis (SA) \cite{pang2008opinion}. SA methods infer the opinion meaning of a fragment of text, and that opinion meaning may be expressed in a binary (positive and negative) or a multi-level (from 1 to 5 stars) scale of opinion intensity. Likewise, SA methods can be conducted at different granularity levels, hence they can be used at document, sentence and aspect level. SA at aspect level is known as aspect-based sentiment analysis (ABSA), and it calculates the opinion meaning of every entity and aspects of those entities explicitly or implicitly mentioned in the text. Hence, SA methods, and more specifically ABSA methods, may be used to overcome the constraint of processing the evaluation of experts in natural language.}


\nuevo{Since SA methods may process the opinion of experts, some works followed a lexicon-based SA approach to infer the position of the experts with respect to a set of target alternatives \cite{morente2018analysing,MORENTEMOLINERA2019335}. However, these lexicon-based SA methods only measure the opinion of the expert at global or evaluation level, they are not able to identify the opinion of the experts when there are several criteria in the DM problem, and they are limited for the lexical coverage of the lexicon.}





\nuevo{We propose in this paper a new methodology for MpMcDM problems that combines SA and DM methods for processing the evaluation of the experts in natural language \segrev{with the aim of providing smarter decision aid.} We call it as Sentiment Analysis based Multi-person Multi-criteria Decision Making (SA-MpMcDM) methodology.} \nuevo{It} 
allow\cz{s} to \cz{deal} with expert evaluations expressed \nuevo{in natural language and even with numerical values.} Accordingly, \nuevo{it} defines how to represent the different kind of evaluations of each expert, and how to combine them for building the input of an \nuevo{MpMcDM} model. Hence, the \nuevo{SA-MpMcDM} methodology allows to use whatever SA and DM model in order to resolve an \nuevo{MpMcDM} problem.

There are several experts and evaluation criteria in \nuevo{MpMcDM} problems, which means that more than one expert will evaluate the alternatives according to different criteria. The criteria of alternatives in a DM problem are different aspects of the alternative susceptible of being assessed, roughly speaking, the alternatives are evaluated at criterion level. \nuevo{As we indicated above, SA and ABSA methods can infer the position of each expert with respect to each criterion. Accordingly, if we consider the criteria of an alternative as aspects of an entity, we can use ABSA methods for processing the expert evaluations about different criteria.}

\nuevo{We propose in this paper the SA-MpMcDM methodology on a three-tier workflow, namely:
\begin{enumerate}
\item Obtaining input evaluations: it compiles the evaluation of the experts about a set of target alternatives in natural language, and if it is available in a numerical scale too. 
\item Distilling opinions at criterion level: the expert evaluations \segrev{expressed} in natural language are processed for distilling their opinions about the target criteria of each alternative. We propose an end-to-end multi-task deep learning model, which we call \nuevo{distilling opinions and criteria using an ABSA based deep learning (DOC-ABSADeepL) model.} It has the capacity of \begin{enumerate*}[label=(\arabic*)] \item identifying all the aspect terms mentioned in the expert evaluation; \item categorizing those aspect terms in aspect categories that correspond to the criteria of an MpMcDM problem; and \item inferring the opinion value about each aspect category.\end{enumerate*} Transforming the opinion values to numbers we obtain a matrix representation of the natural language evaluations. If available, the expert evaluations in numerical scale are also considered. 
\item Alternative choice decisions: the expert evaluations are aggregated into a collective preference matrix, which allows to rank the target alternatives. The aggregation is conducted by \new{the proposed procedure named }
criteria weighting through the attention of experts, \textit{i.e} those criteria that receive more opinions, they more highly attract the attention of the experts, hence the SA-MpMcDM methodology gives them a higher weight in the aggregation.

\end{enumerate}
}






\segrev{We evaluate the SA-MpMcDM methodology in an MpMcDM case study of restaurant choice using restaurant reviews from an e-commerce site.} The restaurants and their services categories are the alternatives and the criteria of the \nuevo{MpMcDM} problem. We compiled and annotated a restaurant review dataset from TripAdvisor\footnote{\url{https://www.tripadvisor.com/}} for evaluating \nuevo{the SA-MpMcDM methodology}, and we call it TripR-2020 dataset.\footnote{The dataset is available at: \url{https://github.com/ari-dasci/OD-TripR-2020}} Each alternative or restaurant is evaluated for all the experts in TripR-2020, which means that can be used for the evaluation of \nuevo{MpMcDM} models.

\nuevo{The main feature of the SA-MpMcDM methodology is enlarging the capacity of \new{an} MpMcDM model of considering evaluative information, by means of leveraging natural language evaluations. Hence, we provide different scenarios based on the methodology 
using \begin{enumerate*}[label=(\arabic*)]\item numerical evaluations and \item natural language evaluations ones.\end{enumerate*} The results show that combining natural language and numerical evaluations enhances the quality of the  MpMcDM model.}


The main contributions of the paper are:
\begin{enumerate}
\item To propose the SA-MpMcDM methodology \segrev{for smarter decision aid}, which combines natural language and numerical expert evaluations in MpMcDM problems.
\item To propose \nuevo{the DOC-ABSADeepL model} that is an end-to-end multi-task deep learning model for ABSA, which allows to \nuevo{process the expert evaluations as opinions at aspect level}.
\item To present the TripR-2020 dataset, which is composed of \segrev{TripAdvisor reviews of restaurants.} 
TripR-2020 can be used for evaluating \nuevo{MpMcDM} models.

\end{enumerate}


This paper is structured as follows. Section \ref{background} presents the bases of DM, ABSA and deep learning to understand the SA-MpMcDM methodology. Section \ref{DMmodel} explains the \nuevo{SA-MpMcDM methodology and its workflow architecture. Section \ref{case_study} describes how the proposed methodology is used in \new{an} MpMcDM problem \segrev{dealing with TripAdvisor restaurant reviews.}} \nuevo{An analysis of the behaviour of the methodology through different scenarios} is shown in Section \ref{com_and_dis}. Finally, conclusions and future work are \nuevo{pointed out} in Section \ref{conclusions}. \new{The \ref{appendixA} summarizes the abbreviations and notations used in this paper.}

\section{Background}\label{background}

The \nuevo{SA-MpMcDM methodology is an MpMcDM model } 
that uses SA methods for understanding the natural language evaluations from the experts. Hence, \nuevo{this} methodology combines several computational methods that may be introduced for its understanding. Accordingly, we present the \nuevo{MpMcDM} task in Section \ref{decision_making} and ABSA in Section \ref{sentiment_analysis}. We expose some works that attain to combine DM and SA in Section \ref{dm+sa}. 
\new{We implement our methodology using deep learning methods, since they have shown state-of-the-art results for SA \cite{zhang2018}, hence we introduce some background knowledge about it in Section \ref{neural_networks}.}


\subsection{Multi-person Multi-criteria Decision Making}\label{decision_making}

DM is a process in which the suitable alternative from a set of possible \geni{ones} \cz{is }
chosen \nuevo{to establish} a ranking of alternatives \cite{triantaphyllou2013multi}. Five steps are usually followed to conduct this process: \begin{enumerate*}[label=(\arabic*)] \item \nuevo{to} identify the problem as well as the objective and the alternatives; \item \nuevo{to} establish a framework that collects all the characteristics of the problem; \item \nuevo{to} obtain the knowledge of the experts who evaluate; \item \nuevo{to} analyze and aggregate the information obtained; and \item \nuevo{to} select the best alternative \cite{fases}\end{enumerate*}. Hence, \geni{a DM process needs to define} a set of alternatives $X=\{x_{1},…,x_{n}\}$ \geni{that} are evaluated by a set of experts $E=\{e_{1},…,e_{l}\}$. The alternatives are usually evaluated according to a set of criteria  $C=\{c_{1},…,c_{m}\}$. These models are known as \nuevo{Multi-person Multi-criteria Decision Making (MpMcDM) models} \nuevo{\cite{ thiriez1976multi, yager1993non}}. 

\begin{figure}[h!]
\tikzset{every picture/.style={line width=0.3pt}} 
\resizebox{0.85\textwidth}{2.7cm}{
\hspace{2.2cm}    

\begin{tikzpicture}[x=0.75pt,y=0.75pt,yscale=-1,xscale=1]

\draw   (21.91,97.21) -- (61.53,97.21) -- (61.53,136.83) -- (21.91,136.83) -- cycle ;
\draw    (22.51,111.35) -- (62.17,111.86) ;
\draw    (22.08,124.09) -- (61.74,124.6) ;
\draw    (35.25,97.08) -- (35.25,136.15) ;
\draw    (47.99,96.65) -- (47.99,137) ;
\draw    (61.5,104.86) -- (70.17,104.86) ;
\draw    (61.5,117.6) -- (69.74,117.6) ;
\draw    (43.25,90.08) -- (43.25,96.2) ;
\draw    (55.99,89.65) -- (55.99,96.2) ;
\draw    (31.5,89.2) -- (70.5,89.2) ;
\draw    (70.5,89.2) -- (70.5,129.2) ;
\draw    (31.5,89.2) -- (31.5,96.2) ;
\draw    (70.5,129.2) -- (61.5,129.2) ;
\draw   (147.91,113.21) -- (187.53,113.21) -- (187.53,152.83) -- (147.91,152.83) -- cycle ;
\draw    (148.51,127.35) -- (188.17,127.86) ;
\draw    (148.08,140.09) -- (187.74,140.6) ;
\draw    (161.25,113.08) -- (161.25,152.15) ;
\draw    (173.99,112.65) -- (173.99,153) ;
\draw    (269.51,116.35) -- (309.17,116.86) ;
\draw    (269.08,129.09) -- (308.74,129.6) ;
\draw    (282.25,115.6) -- (282.25,128.6) ;
\draw    (294.99,115.6) -- (294.99,130) ;
\draw    (309.17,116.86) -- (309.17,129.6) ;
\draw    (269.51,116.35) -- (269.51,129.09) ;
\draw   (126.5,39.6) -- (218.5,39.6) -- (218.5,161.6) -- (126.5,161.6) -- cycle ;
\draw   (245,57.6) -- (330.5,57.6) -- (330.5,143.6) -- (245,143.6) -- cycle ;
\draw   (8.12,43.02) -- (83.85,43.02) -- (83.85,143.87) -- (8.12,143.87) -- cycle ;
\draw    (347,88.6) -- (373.9,88.6) ;
\draw [shift={(375.9,88.6)}, rotate = 180] [color={rgb, 255:red, 0; green, 0; blue, 0 }  ][line width=0.75]    (10.93,-3.29) .. controls (6.95,-1.4) and (3.31,-0.3) .. (0,0) .. controls (3.31,0.3) and (6.95,1.4) .. (10.93,3.29)   ;
\draw    (218,100.6) -- (241.9,100.6) ;
\draw [shift={(243.9,100.6)}, rotate = 180] [color={rgb, 255:red, 0; green, 0; blue, 0 }  ][line width=0.75]    (10.93,-3.29) .. controls (6.95,-1.4) and (3.31,-0.3) .. (0,0) .. controls (3.31,0.3) and (6.95,1.4) .. (10.93,3.29)   ;
\draw    (83,92.6) -- (106.9,92.6) ;
\draw [shift={(108.9,92.6)}, rotate = 180] [color={rgb, 255:red, 0; green, 0; blue, 0 }  ][line width=0.75]    (10.93,-3.29) .. controls (6.95,-1.4) and (3.31,-0.3) .. (0,0) .. controls (3.31,0.3) and (6.95,1.4) .. (10.93,3.29)   ;
\draw   (347.51,170.2) -- (108.74,170.2) -- (108.74,7.2) -- (347.51,7.2) -- cycle ;

\draw (39,71) node [anchor=north west][inner sep=0.75pt]    {$IP$};
\draw (156,91) node [anchor=north west][inner sep=0.75pt]    {$CP$};
\draw (279,97) node [anchor=north west][inner sep=0.75pt]    {$FP$};
\draw (10.12,46.02) node [anchor=north west][inner sep=0.75pt]   [align=left] {\textbf{Input data}};
\draw (246,65.6) node [anchor=north west][inner sep=0.75pt]   [align=left] {Exploitation};
\draw (130,41.6) node [anchor=north west][inner sep=0.75pt]   [align=left] {\begin{minipage}[lt]{56.60224800000001pt}\setlength\topsep{0pt}
\begin{center}
Collective\\aggregation
\end{center}

\end{minipage}};
\draw (376,71.6) node [anchor=north west][inner sep=0.75pt]   [align=left] {\begin{minipage}[lt]{39.563624000000004pt}\setlength\topsep{0pt}
\begin{center}
\textbf{Final }\\\textbf{ranking}
\end{center}

\end{minipage}};
\draw (127,16) node [anchor=north west][inner sep=0.75pt]   [align=left] {\textbf{Selecting the best alternative}};

\end{tikzpicture}
}

\caption{Workflow of a traditional \nuevo{MpMcDM} model.}
\label{worklowDM}
\end{figure}
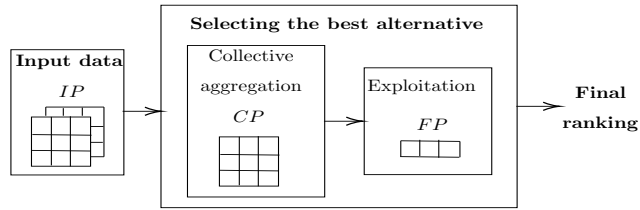

\cz{Figure \ref{worklowDM} shows the workflow of an \nuevo{MpMcDM} process} and can be summarized as follows:
\begin{itemize}[noitemsep]
    \item Input data phase: Experts provide their evaluations about the alternatives. The individual evaluations are collected into matrices.
    \item Selecting the best alternative phase: The evaluations of all the experts are aggregated getting a collective preference matrix. The collective evaluation allows to obtain a ranking of the alternatives by means of an exploitation step\geni{, which gets} the final preference vector.
\end{itemize}



Traditional \nuevo{MpMcDM} models allow experts to provide their individual evaluations through numerical ratings \cite{roubens1997fuzzy} or linguistic variables \nuevo{\cite{herrera2000linguistic, 2tuple, ZADEH1975301}}. The linguistic approaches for computing with words can be classified into \geni{membership functions based models \cite{YAGER1977375, ZadehFuzzy} and qualitative scales based ones \cite{Zuhe1}.} \geni{There are some models that also} allow to aggregate numerical and linguistic information \cite{Herreracombining}. In any case, these models \geni{are constrained by a} pre-defined set of linguistic terms or numerical values \geni{to evaluate} the alternatives.



\subsection{Aspect based Sentiment Analysis}\label{sentiment_analysis}


\nuevo{ABSA is the NLP task mainly focused on the classification of the opinions at aspect level, but it is also concerned with the identification of additional elements of an opinion. According to \cite{liu2015sentiment}, an opinion is defined as the quintuple $(e_{i}, a_{ij}, p_{ijkl}, h_{k}, t_{l})$, such that $e_{i}$ is the name of the evaluated entity, $a_{ij}$ is the aspect term of the entity $e_{i}$, $p_{ijkl}$ is the polarity value of the opinion, $h_{k}$ is the opinion holder and $t_{l}$ is the time when the opinion was written. Therefore, the end-goal of ABSA is the generation of that opinion quintuple.}

\new{In the context of the SA-MpMcDM methodology, the target opinion elements are the polarity of the opinion of each aspect of each entity expressed by each expert. We do not to extract neither the author nor the time in which the raw text was written, so $p_{ijkl}$ is reduced to $p_{ij}$.} 
Hence, the resultant representation of an opinion in the SA-MpMcDM methodology is the tuple $(e_{i}, a_{ij}, p_{ij})$.

\nuevo{The SA-MpMcDM methodology tackles three tasks of ABSA. Given the restaurant review ``I liked the turkey and the vegetables but I didn't like the fruit'', those three tasks are:

\begin{itemize}[noitemsep]
    \item Aspect Terms Extraction:  it is  focused on the identification of all the aspect terms. The aspect terms of the example are: ``turkey'', ``vegetables'' and ``fruit''.
    \item Aspect Category Detection: it categorises the aspect terms in aspect categories. The aspect category of the aspect terms of the example is ``food''.
    \item Classification of the polarity: it classifies the opinion meaning of the opinions about the aspect terms. In the example, the aspect terms ``turkey'' and ``vegetables'' receive a positive opinion, otherwise the aspect term ``fruit'' receives a negative opinion. 
\end{itemize}
}



We develop an ABSA \czm{end-to-end deep learning model that simultaneously performs the three tasks following a multi-task approach.}

\subsection{Sentiment-Analysis-based Decision Making}\label{dm+sa}

\new{There are some works that use SA methods to interpret the evaluations of the experts in a DM setting.} 
These \geni{few} works use lexicon-based SA methods \cite{liu2015sentiment} 
for calculating the opinion value of the evaluation of the experts. These methods are grounded in the use of a list of opinion bearing words, hence the calculation of the opinion value depends on finding out in the evaluations the opinion words of the lexicon. \geni{For instance, the authors of \cite{morente2018analysing,MORENTEMOLINERA2019335} use sentiment lexicons to infer the opinion meaning of the evaluations of the experts. However, these two proposals are constrained by the lexical coverage of the sentiment lexicon, and they lack of a semantic processing of the evaluation of the experts.}


We find other articles that \geni{attain to} connect the two areas but do not provide adequate formalization of DM processes. Actually, they focus on presenting SA \cz{models }
that could be linked to DM \nuevo{\cite{KAUFFMANN2019, 6705664}}. 
These articles lack of: 
\begin{enumerate*}[label=(\arabic*)] 
\item an adequate formalization of the decision-making processes;
\item semantic understanding processes to extract the knowledge of the experts from the provided written texts; and
\item real case studies to test the developed models.
\end{enumerate*}

\subsection{Deep Learning models}\label{neural_networks}


\geni{Deep learning models have been widely used in NLP \nuevo{\cite{goldberg2017neural}}. This section presents the basis of convolutional neural networks (CNN) and long-short-term memories (LSTM) networks, since they are used in the developed ABSA model. Furthermore, multi-task \cris{deep learning is} also introduced.}

\subsubsection{Convolutional Neural Networks}

CNN aim to identify local predictors in large structures with grid-like topologies, \textit{i.e.}, CNN layers try to extract meaningful sub-structures which are representative for the prediction. These networks are characterized by using the mathematical operation known as convolution, which is a kind of linear operation. This operation receives two arguments, the input and the kernel, and its output is known as the feature map \cite{Goodfellow-et-al-2016}.

\geni{The performance of deep learning models based on CNN} can be improved through integrating the  attention mechanism \cite{bahdanau2014neural}. It allows to capture the correlation between non-consecutive words focusing the attention on the specific significant words. The attention mechanism adds a fully-connected layer able to learn the importance and relations between words. CNNs and CNNs with attention have been applied on NLP tasks achieving a strong performance \cite{jacovi2018understanding, yin2015abcnn}.


\subsubsection{Long Short-Term Memory}
Recurrent Neural Networks (RNNs) are defined by means of a function $R$ that is recursively applied on a sequence of input words $(w_{1},w_{2},...,w_{s})$. Specifically, the $R$ function consider\cris{s} as input a vector $w_{i}$ and a state vector $s_{i-1}$, producing as output a new state vector named $s_{i}$. The new state vector is applied to a deterministic function $O(\cdot)$ getting an output vector $y_{i}$. The formal definition is captured in Equation \ref{eq_rnn_definition} \cite{goldberg2017neural}.
\begin{equation}
\begin{split}
RNN(w_{1:s};\textbf{s}_{o})		&=	\textbf{y}_{1:s}\\
				\textbf{y}_{i}	&=	O(\textbf{s}_{i})\\
                \textbf{s}_{i}	&=	R(\textbf{s}_{i-1},\textbf{w}_i);\\
\textbf{w}_i \in \mathds{R}^{d}, \; \; \; \textbf{y}_{i} & \in \mathds{R}^{out}, \; \; \; \textbf{s}_{i} \in \mathds{R}^{f(out)}
\end{split}
\label{eq_rnn_definition}
\end{equation}

Long short-term memory (LSTM) is a kind of RNN \cite{10.1162/neco.1997.9.8.1735}. The main contribution of LSTM is to solve the vanishing gradient problem of RNNs \cite{pascanu2012difficulty}. To achieve it, the gating-based architecture of the LSTM analyzes how much of the input word should be kept or forgotten \cite{10.1162/089976600300015015}. Then, each input word is encoded with the meaning of the significant previous words.

From a linguistic point of view, the meaning of a word is affected not only by its previous words but also by the following words. Therefore, the bidirectional LSTM (biLSTM) arises. This network allows to encode all the information through two consecutive LSTM networks, one encoding forward information (LSTM$^{f}$) and one encoding backward information (LSTM$^{b}$). Equation \ref{eq_lstm} summarises the definition:
\begin{equation}
\begin{split}
biLSTM(w_{1:s})	&=	[LSTM^{f}(w_{1:s},\textbf{s}^{f}_{o});LSTM^{b}(w_{1:s},\textbf{s}^{b}_{o})]
\end{split}
\label{eq_lstm}
\end{equation}

LSTM and biLSTM are currently the most successful networks in NLP. They are used in different tasks such as automatic language identification \cite{GonzalezDominguez2014AutomaticLI}, word sense disambiguation \cite{Zuhe2} \geni{and ABSA \cite{Shuang2020}.}

\subsubsection{Multi-task deep learning}
\cris{Multi-Task Learning (MTL)} is a sub-task of machine learning which leverages information from multiple related tasks to improve the overall performance of all the tasks \cite{zhang2017survey}. Learning each task helps rest of tasks to be learned better by means of the back-propagation mechanism\geni{, which} allows features obtained in the hidden layers for one task to be used in other tasks \cite{caruana1997multitask}.

\begin{figure}[!h]
\centering
\tikzset{every picture/.style={line width=0.2pt}} 
\resizebox{0.7\textwidth}{1.5cm}{

\begin{tikzpicture}[x=0.75pt,y=0.75pt,yscale=-1,xscale=1]

\draw   (66,37) -- (168.5,37) -- (168.5,52.2) -- (66,52.2) -- cycle ;
\draw   (207,22) -- (309.5,22) -- (309.5,37.2) -- (207,37.2) -- cycle ;
\draw   (208,52) -- (310.5,52) -- (310.5,67.2) -- (208,67.2) -- cycle ;
\draw    (45,45) -- (62.5,45) ;
\draw [shift={(64.5,45)}, rotate = 180] [color={rgb, 255:red, 0; green, 0; blue, 0 }  ][line width=0.75]    (10.93,-3.29) .. controls (6.95,-1.4) and (3.31,-0.3) .. (0,0) .. controls (3.31,0.3) and (6.95,1.4) .. (10.93,3.29)   ;
\draw    (319,29) -- (336.5,29) ;
\draw [shift={(338.5,29)}, rotate = 180] [color={rgb, 255:red, 0; green, 0; blue, 0 }  ][line width=0.75]    (10.93,-3.29) .. controls (6.95,-1.4) and (3.31,-0.3) .. (0,0) .. controls (3.31,0.3) and (6.95,1.4) .. (10.93,3.29)   ;
\draw    (319,58) -- (336.5,58) ;
\draw [shift={(338.5,58)}, rotate = 180] [color={rgb, 255:red, 0; green, 0; blue, 0 }  ][line width=0.75]    (10.93,-3.29) .. controls (6.95,-1.4) and (3.31,-0.3) .. (0,0) .. controls (3.31,0.3) and (6.95,1.4) .. (10.93,3.29)   ;
\draw    (178,51) -- (193.73,59.27) ;
\draw [shift={(195.5,60.2)}, rotate = 207.73] [color={rgb, 255:red, 0; green, 0; blue, 0 }  ][line width=0.75]    (10.93,-3.29) .. controls (6.95,-1.4) and (3.31,-0.3) .. (0,0) .. controls (3.31,0.3) and (6.95,1.4) .. (10.93,3.29)   ;
\draw    (176,39) -- (194.7,30.06) ;
\draw [shift={(196.5,29.2)}, rotate = 514.45] [color={rgb, 255:red, 0; green, 0; blue, 0 }  ][line width=0.75]    (10.93,-3.29) .. controls (6.95,-1.4) and (3.31,-0.3) .. (0,0) .. controls (3.31,0.3) and (6.95,1.4) .. (10.93,3.29)   ;

\draw (73,3) node [anchor=north west][inner sep=0.75pt]   [align=left] {shared layers};
\draw (195,2) node [anchor=north west][inner sep=0.75pt]   [align=left] {task especific layers};
\draw (346,19.4) node [anchor=north west][inner sep=0.75pt]    {$task\ 1$};
\draw (344,48.4) node [anchor=north west][inner sep=0.75pt]    {$task\ n$};
\draw (4,34.4) node [anchor=north west][inner sep=0.75pt]    {$input$};
\draw (354,39.4) node [anchor=north west][inner sep=0.75pt]    {$...$};

\end{tikzpicture}
}
\caption{Multi-task \cris{deep learning architecture} with hard parameter sharing.}
\label{squeme_multitask}
\end{figure}
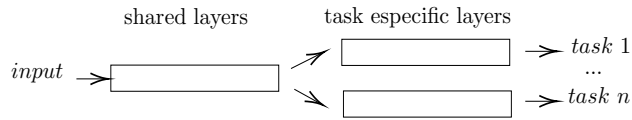

There are two sort of structures to perform multi-task learning through neural networks depending on the parameter sharing of the hidden layers:  
\begin{enumerate*}[label=(\arabic*)] 
\item soft parameter sharing architectures, in which each task has its own parameters that maintain a distance that is regularized \cite{duong-etal-2015-low}; and
\item hard parameter sharing architectures that share hidden layers between all tasks and maintain specific output layers for each task \cite{Caruana93multitasklearning}.
\end{enumerate*} This last type of structure is most commonly used since it avoids overfitting. Figure \ref{squeme_multitask} \geni{outlines} the hard parameter sharing architecture.

Multi-task \cris{deep learning} are successfully used in different NLP tasks, such as question answering \cite{mccann2018natural} and \czm{language understanding \cite{liu2019multitask}.}

\section{Sentiment Analysis based Multi-person Multi-criteria Decision Making: SA-MpMcDM Methodology}\label{DMmodel}

\nuevo{MpMcDM} models allow experts to provide their evaluations through numerical evaluations or linguistics terms \cite{Herreracombining}. Evaluating through numerical values causes difficulties for experts, since it is not natural for humans to express their opinions \eugenio{with} numbers. Alternatively, evaluating using linguistic terms means that expert evaluations are limited to those terms. In any case, those models cause that the evaluations of the experts are not as valuable and right as they could and should be. 




We propose the \nuevo{SA-MpMcDM} methodology that allows experts to freely evaluate the alternatives using natural language, and more specifically experts are able to evaluate the alternatives in multiple ways, for instance: \begin{enumerate*}[label=(\arabic*)] \item publishing reviews on \nuevo{e-commerce} sites; \item filling out questionnaires using free text; \item publishing posts about products; \item and even answering surveys orally.\end{enumerate*}

In this paper, we focus on \nuevo{the first of the four options, \textit{i.e.,} on }e-commerce sites. \nuevo{Consequently, }the experts are the users that publish reviews about different entities, which are the alternatives to evaluate. 

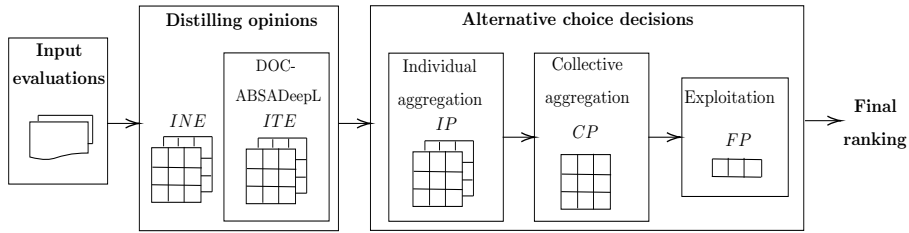
\begin{figure}[!h]
\tikzset{every picture/.style={line width=0.3pt}} 
\resizebox{\textwidth}{3cm}{ 

\begin{tikzpicture}[x=0.75pt,y=0.75pt,yscale=-1,xscale=1]

\draw   (18,91.52) -- (64.82,91.52) -- (64.82,114.09) .. controls (35.56,114.09) and (41.41,122.23) .. (18,116.96) -- cycle ;
\draw    (24.69,86.73) -- (68.5,86.73) ;
\draw    (68.5,86.73) -- (68.5,107.25) ;
\draw    (24.69,86.73) -- (24.69,90.84) ;
\draw    (68.5,107.25) -- (64.49,107.25) ;
\draw   (191.91,111.21) -- (231.53,111.21) -- (231.53,150.83) -- (191.91,150.83) -- cycle ;
\draw    (192.51,125.35) -- (232.17,125.86) ;
\draw    (192.08,138.09) -- (231.74,138.6) ;
\draw    (205.25,111.08) -- (205.25,150.15) ;
\draw    (217.99,110.65) -- (217.99,151) ;
\draw    (231.5,118.86) -- (240.17,118.86) ;
\draw    (231.5,131.6) -- (239.74,131.6) ;
\draw    (213.25,104.08) -- (213.25,110.2) ;
\draw    (225.99,103.65) -- (225.99,110.2) ;
\draw    (201.5,103.2) -- (240.5,103.2) ;
\draw    (240.5,103.2) -- (240.5,143.2) ;
\draw    (201.5,103.2) -- (201.5,110.2) ;
\draw    (240.5,143.2) -- (231.5,143.2) ;
\draw   (115.91,110.21) -- (155.53,110.21) -- (155.53,149.83) -- (115.91,149.83) -- cycle ;
\draw    (116.51,124.35) -- (156.17,124.86) ;
\draw    (116.08,137.09) -- (155.74,137.6) ;
\draw    (129.25,110.08) -- (129.25,149.15) ;
\draw    (141.99,109.65) -- (141.99,150) ;
\draw    (155.5,117.86) -- (164.17,117.86) ;
\draw    (155.5,130.6) -- (163.74,130.6) ;
\draw    (137.25,103.08) -- (137.25,109.2) ;
\draw    (149.99,102.65) -- (149.99,109.2) ;
\draw    (125.5,102.2) -- (164.5,102.2) ;
\draw    (164.5,102.2) -- (164.5,142.2) ;
\draw    (125.5,102.2) -- (125.5,109.2) ;
\draw    (164.5,142.2) -- (155.5,142.2) ;
\draw   (325.91,113.21) -- (365.53,113.21) -- (365.53,152.83) -- (325.91,152.83) -- cycle ;
\draw    (326.51,127.35) -- (366.17,127.86) ;
\draw    (326.08,140.09) -- (365.74,140.6) ;
\draw    (339.25,113.08) -- (339.25,152.15) ;
\draw    (351.99,112.65) -- (351.99,153) ;
\draw    (365.5,120.86) -- (374.17,120.86) ;
\draw    (365.5,133.6) -- (373.74,133.6) ;
\draw    (347.25,106.08) -- (347.25,112.2) ;
\draw    (359.99,105.65) -- (359.99,112.2) ;
\draw    (335.5,105.2) -- (374.5,105.2) ;
\draw    (374.5,105.2) -- (374.5,145.2) ;
\draw    (335.5,105.2) -- (335.5,112.2) ;
\draw    (374.5,145.2) -- (365.5,145.2) ;
\draw   (444.91,115.21) -- (484.53,115.21) -- (484.53,154.83) -- (444.91,154.83) -- cycle ;
\draw    (445.51,129.35) -- (485.17,129.86) ;
\draw    (445.08,142.09) -- (484.74,142.6) ;
\draw    (458.25,115.08) -- (458.25,154.15) ;
\draw    (470.99,114.65) -- (470.99,155) ;
\draw    (566.51,118.35) -- (606.17,118.86) ;
\draw    (566.08,131.09) -- (605.74,131.6) ;
\draw    (579.25,117.6) -- (579.25,130.6) ;
\draw    (591.99,117.6) -- (591.99,132) ;
\draw    (606.17,118.86) -- (606.17,131.6) ;
\draw    (566.51,118.35) -- (566.51,131.09) ;
\draw   (106,7.42) -- (266,7.42) -- (266,170.42) -- (106,170.42) -- cycle ;
\draw   (306.5,41.6) -- (398.5,41.6) -- (398.5,163.6) -- (306.5,163.6) -- cycle ;
\draw   (423.5,41.6) -- (515.5,41.6) -- (515.5,163.6) -- (423.5,163.6) -- cycle ;
\draw   (542,59.6) -- (627.5,59.6) -- (627.5,145.6) -- (542,145.6) -- cycle ;
\draw   (1,30.6) -- (80.5,30.6) -- (80.5,136.6) -- (1,136.6) -- cycle ;
\draw    (641,90.6) -- (667.9,90.6) ;
\draw [shift={(669.9,90.6)}, rotate = 180] [color={rgb, 255:red, 0; green, 0; blue, 0 }  ][line width=0.75]    (10.93,-3.29) .. controls (6.95,-1.4) and (3.31,-0.3) .. (0,0) .. controls (3.31,0.3) and (6.95,1.4) .. (10.93,3.29)   ;
\draw    (515,102.6) -- (538.9,102.6) ;
\draw [shift={(540.9,102.6)}, rotate = 180] [color={rgb, 255:red, 0; green, 0; blue, 0 }  ][line width=0.75]    (10.93,-3.29) .. controls (6.95,-1.4) and (3.31,-0.3) .. (0,0) .. controls (3.31,0.3) and (6.95,1.4) .. (10.93,3.29)   ;
\draw    (398,102.6) -- (421.9,102.6) ;
\draw [shift={(423.9,102.6)}, rotate = 180] [color={rgb, 255:red, 0; green, 0; blue, 0 }  ][line width=0.75]    (10.93,-3.29) .. controls (6.95,-1.4) and (3.31,-0.3) .. (0,0) .. controls (3.31,0.3) and (6.95,1.4) .. (10.93,3.29)   ;
\draw    (266,92.6) -- (289.9,92.6) ;
\draw [shift={(291.9,92.6)}, rotate = 180] [color={rgb, 255:red, 0; green, 0; blue, 0 }  ][line width=0.75]    (10.93,-3.29) .. controls (6.95,-1.4) and (3.31,-0.3) .. (0,0) .. controls (3.31,0.3) and (6.95,1.4) .. (10.93,3.29)   ;
\draw    (80,92.6) -- (103.9,92.6) ;
\draw [shift={(105.9,92.6)}, rotate = 180] [color={rgb, 255:red, 0; green, 0; blue, 0 }  ][line width=0.75]    (10.93,-3.29) .. controls (6.95,-1.4) and (3.31,-0.3) .. (0,0) .. controls (3.31,0.3) and (6.95,1.4) .. (10.93,3.29)   ;
\draw   (639.97,170.2) -- (291.74,170.2) -- (291.74,7.2) -- (639.97,7.2) -- cycle ;
\draw   (173.99,41.42) -- (258.51,41.42) -- (258.51,163.42) -- (173.99,163.42) -- cycle ;

\draw (126,11.6) node [anchor=north west][inner sep=0.75pt]   [align=left] {\textbf{Distilling opinions}};
\draw (204,85) node [anchor=north west][inner sep=0.75pt]    {$ITE$};
\draw (129,85) node [anchor=north west][inner sep=0.75pt]    {$INE$};
\draw (343,87) node [anchor=north west][inner sep=0.75pt]    {$IP$};
\draw (453,93) node [anchor=north west][inner sep=0.75pt]    {$CP$};
\draw (576,99) node [anchor=north west][inner sep=0.75pt]    {$FP$};
\draw (22,33.6) node [anchor=north west][inner sep=0.75pt]   [align=left] {\textbf{Input}};
\draw (3,53.6) node [anchor=north west][inner sep=0.75pt]   [align=left] {\textbf{evaluations}};
\draw (543,67.6) node [anchor=north west][inner sep=0.75pt]   [align=left] {Exploitation};
\draw (427,43.6) node [anchor=north west][inner sep=0.75pt]   [align=left] {\begin{minipage}[lt]{56.60224800000001pt}\setlength\topsep{0pt}
\begin{center}
Collective\\aggregation
\end{center}

\end{minipage}};
\draw (308.5,44.6) node [anchor=north west][inner sep=0.75pt]   [align=left] {\begin{minipage}[lt]{56.60224800000001pt}\setlength\topsep{0pt}
\begin{center}
Individual\\aggregation
\end{center}

\end{minipage}};
\draw (670,73.6) node [anchor=north west][inner sep=0.75pt]   [align=left] {\begin{minipage}[lt]{39.563624000000004pt}\setlength\topsep{0pt}
\begin{center}
\textbf{Final }\\\textbf{ranking}
\end{center}

\end{minipage}};
\draw (364,11) node [anchor=north west][inner sep=0.75pt]   [align=left] {\textbf{Alternative choice decisions}};
\draw (182,45) node [anchor=north west][inner sep=0.75pt]   [align=left] {\begin{minipage}[lt]{48.643188pt}\setlength\topsep{0pt}
\begin{center}
\small{DOC-\\ABSADeepL}
\end{center}
\end{minipage}};


\end{tikzpicture}

}
\caption{Workflow of the \nuevo{SA-MpMcDM} methodology.}
\label{worklowDM_SA}
\end{figure}

\geni{In contrast to} the traditional scheme of \nuevo{MpMcDM} models shown in Figure \ref{worklowDM}, \nuevo{SA-MpMcDM} incorporates a phase in which the expert knowledge is extracted by means of a semantic understanding procedure. Specifically, \cz{Figure \ref{worklowDM_SA} presents the \nuevo{SA-MpMcDM} workflow that is split into three phases:}



\nuevo{
\begin{itemize}[noitemsep]
    \item Obtaining input evaluations phase: the experts evaluations, written texts in natural language and even numerical ratings, are extracted. 
    \item Distilling opinions at criterion level phase: the opinions from the written texts are extracted using the DOC-ABSADeepL model that incorporates \new{an} NLP procedure focus on ABSA and collected into matrices. Optionally, the numerical ratings are collected into matrices too.
    \item Alternative choice decisions phase: the ranking of the alternatives is obtained aggregating the evaluation of all experts through the weighted criteria, which are obtained by the proposed procedure named criteria weighting through the attention of the experts. 
\end{itemize}}

\nuevo{We subsequently describe the three phases, specifically the \textit{obtaining input evaluations phase}} is exposed in Section \ref{input_phase}, the \nuevo{\textit{distilling opinions at criterion level phase}} is discussed in Section \ref{data_preparation_phase} and the \nuevo{\textit{Alternative choice decisions phase}} is explained in Section \ref{selecting_best_phase}. \nuevo{We outline the full architecture of the \nuevo{SA-MpMcDM} methodology in Section \ref{full_arq}. \new{We compile all the notation used in \ref{appendixA}}.}

\subsection{Obtaining input evaluations phase}\label{input_phase}
Given a \nuevo{MpMcDM} problem, the first task is to define the set of alternatives and the set of experts. Since we focus on e-commerce sites, the alternatives \geni{and the experts are selected from the e-commerce site.}

The set of alternatives $X=\{x_{1},…,x_{n}\}$ is the set of entities from the e-commerce site that could solve the \nuevo{MpMcDM} problem. Each alternative can be evaluated according to a set of pre-defined criteria $C=\{c_{1},…,c_{m}\}$. The set of experts $E=\{e_{1},…,e_{l}\}$ is the set of users that evaluate all the alternatives. \nuevo{The methodology} allows to fix this set in two ways: 
\begin{enumerate*}[label=(\arabic*)] 
\item experts could be fixed a priori, so we only consider the reviews that those users provide about the alternatives; and 
\item experts could be fixed a posteriori, so we consider the reviews of all the possible users from the e-commerce site who have evaluated the alternatives. Second \cz{approach} allows to have as many experts as possible \nuevo{given place to large scale decision making \cite{DING202084}}.
\end{enumerate*}

\nuevo{In order to extract the evaluations of the expert to the alternatives from the e-commerce sites, a web crawler is needed. It allows to download from the web site all the information related to the expert, the alternative and the evaluation such as the written texts in natural language and even the numerical ratings.}

\nuevo{The written text evaluations provided by each expert $e_{k}, k=1,...,l$ are collected into a vector $T^{k}$. }
Therefore, each element $t^{k}_{i}$ is the \nuevo{written text in natural language} of the user $e_{k}$ to the alternative $x_{i}, i=1,...,n$. Each written text is made up of a set of sentences, and each sentence can expose multiple opinions. Each opinion expresses an \nuevo{evaluation} about a criterion.

In addition, many e-commerce sites allow users to evaluate entities not only through natural language \czm{written texts, }but assigning numerical ratings to certain criteria \prev{related to} the entities \nuevo{too}. Traditionally, these platforms provide an intensity level opinion system for users to numerically evaluate a criterion. Consider that experts may provide their numerical evaluations in a scale of $2\tau+1$ levels of opinion intensity. \prev{Then, the numerical ratings belong to interval $[-\tau, \tau]$. To achieve unanimity, all the numerical values inferred or analyzed by \nuevo{SA-MpMcDM} belong to this interval.}



\eugenio{To sum up, Figure \ref{squeme_phase1} outlines the \nuevo{\textit{obtaining input evaluation phase}} of} \nuevo{the SA-MpMcDM methodology}. The output of the \textit{input evaluation phase}, \textit{i.e.}, the numerical ratings and the written texts in natural language, is the input of the \nuevo{\textit{distilling opinions at criterion level phase}}.
\input{tikzs/squeme_Phase1.tex}

\subsection{Distilling opinions at criterion level phase}\label{data_preparation_phase}
The \nuevo{\textit{distilling opinions at criterion level phase} collects the written texts and even the numerical ratings} provided by the experts into the \nuevo{individual textual evaluation (\textit{ITE}) and individual numerical evaluation (\textit{INE}) matrices} respectively. \nuevo{The $INE$ matrices are optional since not all the e-commerce site allows users to provide numerical ratings. Furthermore, the $INE$ matrices provide less information than the $ITE$ matrices, since the $INE$ matrices are limited to the criteria that web sites allow to evaluate \new{through} numerical ratings, while the $ITE$ matrices are not limited by pre-defined criteria.} \nuevo{Section \ref{sentiment_analysis_phase} describes the process of filling the $ITE$ matrices, while Section \ref{obtainINE} presents the process of filling the $INE$ matrices with the optional numerical evaluations.} Figure \ref{squeme_phase2} illustrates the \nuevo{distilling opinions at criterion level phase}.
\input{tikzs/squeme_Phase2_v2.tex}

\subsubsection{Individual Textual Evaluations with the DOC-ABSADeepL model}\label{sentiment_analysis_phase}

\eugenio{An individual textual evaluation matrix ($ITE$) is defined for each expert. \nuevo{The rows represent the alternatives while the columns represent the criteria. The matrices are built extracting the opinions \nuevo{using the} DOC-ABSADeepL model, which processes the evaluations of the undefined criteria of each alternative from the written texts, and transforming the opinions into numbers. Therefore, the generation of the $ITE$ matrices is split} into \nuevo{two} subtasks, namely: \begin{enumerate*}[label=(\arabic*)] \item extracting the opinions and \item computing the $ite$ values. \end{enumerate*}}



\eugenio{
\paragraph{Extract the opinions} The target information for building the $ITE$ matrices is the aspects of each entity (\cz{criteria}
) and the value of the opinion about each aspect. Likewise, the aspect terms may refer to similar concepts, hence they may be categorized in aspect categories. For instance, if the review is talking about the \textit{main meal} and the \textit{desert}, those aspect terms can be categorized in the aspect category \textit{food}. Therefore, we need to extract the aspect terms, the aspect categories and the value of the opinion.

}

\czm{
To extract the aspects, the categories and the polarities of the opinions, two approaches can be followed: \begin{enumerate*}[label=(\arabic*)] 
\item an individual approach to independently extract aspects, categories and polarities; or
\item an end-to-end multi-task approach to simultaneously \geni{infer them.}
\end{enumerate*} We focus on the second methodology, since we consider that there is a strong relation between the aspect, the category and the polarity of the opinions. We thus develop a multi-task end-to-end deep learning model, \nuevo{named DOC-ABSADeepL model,} with language understanding capacity in the line of the state of the art in ABSA \cite{heRuidan2019}.


The \nuevo{DOC-ABSADeepL model }analyzes each sentence of the natural language reviews of the experts (elements $t^{k}_{i}$) to identify the opinion values. In particular, Figure \ref{squeme_network} presents the designed model to extract the aspect, the category and the polarity of each identified opinion. 
} The architecture of the \nuevo{DOC-ABSADeepL model}, shown in more detail in Equation \ref{network_layers}, is composed by: 

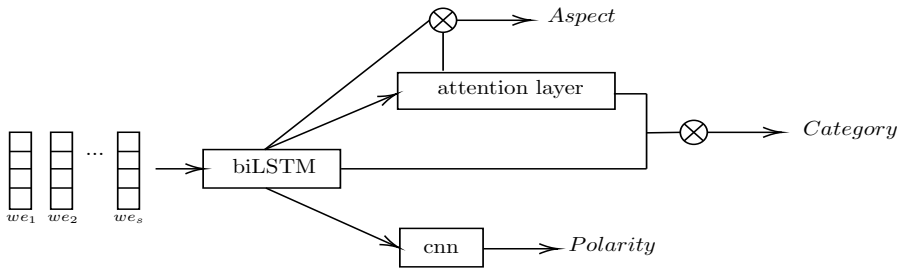
\begin{figure}[h!]
\tikzset{every picture/.style={line width=0.75pt}} 
\resizebox{\textwidth}{3.5cm}{      

\begin{tikzpicture}[x=0.75pt,y=0.75pt,yscale=-1,xscale=1]

\draw   (121.16,99.78) -- (201.09,99.78) -- (201.09,124.95) -- (121.16,124.95) -- cycle ;
\draw   (8.5,88.45) -- (21.09,88.45) -- (21.09,138.8) -- (8.5,138.8) -- cycle ;
\draw    (8.5,101.04) -- (21.09,101.04) ;
\draw    (8.5,124.95) -- (21.09,124.95) ;
\draw    (8.5,112.37) -- (21.09,112.37) ;
\draw   (32.42,88.45) -- (45,88.45) -- (45,138.8) -- (32.42,138.8) -- cycle ;
\draw    (32.42,101.04) -- (45,101.04) ;
\draw    (32.42,124.95) -- (45,124.95) ;
\draw    (32.42,112.37) -- (45,112.37) ;
\draw   (71.44,88.45) -- (84.03,88.45) -- (84.03,138.8) -- (71.44,138.8) -- cycle ;
\draw    (71.44,101.04) -- (84.03,101.04) ;
\draw    (71.44,124.95) -- (84.03,124.95) ;
\draw    (71.44,112.37) -- (84.03,112.37) ;
\draw   (234.45,49.43) -- (360.96,49.43) -- (360.96,73.34) -- (234.45,73.34) -- cycle ;
\draw   (235.71,151.39) -- (284.17,151.39) -- (284.17,176.56) -- (235.71,176.56) -- cycle ;
\draw    (157.67,124.95) -- (233.3,163.08) ;
\draw [shift={(235.08,163.98)}, rotate = 206.75] [color={rgb, 255:red, 0; green, 0; blue, 0 }  ][line width=0.75]    (10.93,-3.29) .. controls (6.95,-1.4) and (3.31,-0.3) .. (0,0) .. controls (3.31,0.3) and (6.95,1.4) .. (10.93,3.29)   ;
\draw    (157.04,99.78) -- (232.02,64.13) ;
\draw [shift={(233.82,63.27)}, rotate = 514.5699999999999] [color={rgb, 255:red, 0; green, 0; blue, 0 }  ][line width=0.75]    (10.93,-3.29) .. controls (6.95,-1.4) and (3.31,-0.3) .. (0,0) .. controls (3.31,0.3) and (6.95,1.4) .. (10.93,3.29)   ;
\draw    (284.39,164.72) -- (325.82,164.72) ;
\draw [shift={(327.82,164.72)}, rotate = 180] [color={rgb, 255:red, 0; green, 0; blue, 0 }  ][line width=0.75]    (10.93,-3.29) .. controls (6.95,-1.4) and (3.31,-0.3) .. (0,0) .. controls (3.31,0.3) and (6.95,1.4) .. (10.93,3.29)   ;
\draw    (360.33,63.27) -- (379.37,63.27) ;
\draw    (201.5,112.37) -- (379.37,112.37) ;
\draw    (379.37,63.27) -- (379.37,112.37) ;
\draw   (399,88.5) .. controls (399,84.16) and (402.52,80.64) .. (406.87,80.64) .. controls (411.21,80.64) and (414.73,84.16) .. (414.73,88.5) .. controls (414.73,92.85) and (411.21,96.37) .. (406.87,96.37) .. controls (402.52,96.37) and (399,92.85) .. (399,88.5) -- cycle ; \draw   (401.3,82.94) -- (412.43,94.07) ; \draw   (412.43,82.94) -- (401.3,94.07) ;
\draw    (379.49,88.82) -- (399,88.5) ;
\draw    (414.73,88.5) -- (456.16,88.5) ;
\draw [shift={(458.16,88.5)}, rotate = 180] [color={rgb, 255:red, 0; green, 0; blue, 0 }  ][line width=0.75]    (10.93,-3.29) .. controls (6.95,-1.4) and (3.31,-0.3) .. (0,0) .. controls (3.31,0.3) and (6.95,1.4) .. (10.93,3.29)   ;
\draw   (253.02,15.12) .. controls (253.02,10.78) and (256.54,7.26) .. (260.89,7.26) .. controls (265.23,7.26) and (268.75,10.78) .. (268.75,15.12) .. controls (268.75,19.47) and (265.23,22.99) .. (260.89,22.99) .. controls (256.54,22.99) and (253.02,19.47) .. (253.02,15.12) -- cycle ; \draw   (255.32,9.56) -- (266.45,20.69) ; \draw   (266.45,9.56) -- (255.32,20.69) ;
\draw    (260.89,22.99) -- (260.89,48.17) ;
\draw    (157.04,99.78) -- (253.02,15.12) ;
\draw    (268.75,15.12) -- (310.18,15.12) ;
\draw [shift={(312.18,15.12)}, rotate = 180] [color={rgb, 255:red, 0; green, 0; blue, 0 }  ][line width=0.75]    (10.93,-3.29) .. controls (6.95,-1.4) and (3.31,-0.3) .. (0,0) .. controls (3.31,0.3) and (6.95,1.4) .. (10.93,3.29)   ;
\draw    (93.47,112.37) -- (119.79,112.37) ;
\draw [shift={(121.79,112.37)}, rotate = 180] [color={rgb, 255:red, 0; green, 0; blue, 0 }  ][line width=0.75]    (10.93,-3.29) .. controls (6.95,-1.4) and (3.31,-0.3) .. (0,0) .. controls (3.31,0.3) and (6.95,1.4) .. (10.93,3.29)   ;

\draw (136.83,104.49) node [anchor=north west][inner sep=0.75pt]   [align=left] {biLSTM};
\draw (256.04,52.88) node [anchor=north west][inner sep=0.75pt]   [align=left] {attention layer};
\draw (247.89,160) node [anchor=north west][inner sep=0.75pt]   [align=left] {cnn};
\draw (320.1,5.32) node [anchor=north west][inner sep=0.75pt]    {$Aspect$};
\draw (468.96,79.33) node [anchor=north west][inner sep=0.75pt]    {$Category$};
\draw (332.34,155.12) node [anchor=north west][inner sep=0.75pt]    {$Polarity$};
\draw (51.22,100.72) node [anchor=north west][inner sep=0.75pt]   [align=left] {...};
\draw (5,140.73) node [anchor=north west][inner sep=0.75pt]  [font=\scriptsize]  {$we_{1}$};
\draw (29,140.73) node [anchor=north west][inner sep=0.75pt]  [font=\scriptsize]  {$we_{2}$};
\draw (67.44,141.2) node [anchor=north west][inner sep=0.75pt]  [font=\scriptsize]  {$we_{s}$};

\end{tikzpicture}

}
\caption{Architecture of the \nuevo{DOC-ABSADeepL model} for extracting the triplet \textit{(aspect, category, polarity)} of all the opinions provided in a sentence.}
\label{squeme_network}
\end{figure}

\begin{itemize}[noitemsep, leftmargin=*]
\item Embedding layer. The input layer is defined by a sequence of $s$ words, $\{w_{1}, . . . ,w_{s}\}$. We \geni{represent each input word ($w_{r}$) with its corresponding $d$-dimensional word embedding vector $\textbf{we}_{r} = (we_{r1},...,we_{rd})$.  The output of the embedding layer is $we_{s\times d}$.} 

\item Shared layer. To capture the dependencies between words, we add a \eugenio{bidirectional LSTM layer (biLSTM)} with $h_{lstm}$ hidden units. The output of the biLSTM layer are features that encode the inter-dependencies among words.


\item Classification layer. \geni{It is composed of three 
layers, one per each task:}
\geni{
\begin{enumerate}

\item Aspect term classification. We combine the outputs of the biLSTM and the attention layers for leveraging the semantic information from the inter-dependencies of the words.  The output of aspect term classification layer is a $s$-dimensional vector that labels whether a word is an aspect.

\item Aspect category classification. It is similar to the previous one, because it also combines the outputs of the biLSTM and the attention layers. The output is a $s$-dimensional vector indicating the aspect category to which each word belongs in case it expresses an opinion.

\item Polarity classification. It combines the outputs of the biLSTM layer and the CNN layer, which has as kernel size $k_{1}$ and a feature maps of size $fm_{1}$. The output is a $s$-dimensional vector that label the opinion \cris{value of} each word regarding the aspect and the identified category.
\end{enumerate}
}
Aspects and categories are obtained by the same process sharing all layers since they are strongly connected. Polarities are obtained by a parallel classification layer since is not as closely dependent on aspects and categories. 

\end{itemize}

\vspace*{-\baselineskip}

\begin{equation}
\begin{split}
\textbf{pred\_aspect} &=  sigmoid(y^{3}_{s\times2h_{lstm}})\\
\textbf{pred\_category} &=  sigmoid(y^{4}_{s\times2h_{lstm}})\\
\textbf{pred\_polarity} &=  sigmoid(y^{5}_{s \times fm_{1}})\\
\textbf{y}^{5}_{s \times fm_{1}}	&= CNN(y^{1}_{s\times2h_{lstm}},k_{1})\\
\textbf{y}^{4}_{s\times2h_{lstm}}	&= multiply(y^{1}_{s\times2h_{lstm}},y^{2}_{s\times2h_{lstm}})\\
\textbf{y}^{3}_{s\times2h_{lstm}}	&= multiply(y^{1}_{s\times2h_{lstm}},y^{2}_{s\times2h_{lstm}})\\
\textbf{y}^{2}_{s\times2h_{lstm}}	&= cnn\_attention(y^{1}_{s\times2h_{lstm}})\\
\textbf{y}^{1}_{s\times2h_{lstm}}	&= biLSTM(we_{s\times d})
\end{split}
\label{network_layers}
\end{equation}

One of the most outstanding advantages of the \nuevo{DOC-ABSADeepL} model is providing mutual information between aspects, categories and polarities. Moreover, aspects and categories are strongly related in the model as in the raw text. For example, the aspect \textit{beer} is \eugenio{usually} related to the \textit{drink} criterion, so when identifying such aspect the model easily associate its criterion.

Another 
advantage of the model is the capacity of obtaining the categories and the polarities of implicit aspects. This is really useful for sentences that provide an overall opinion on a criterion and not a specific opinion on an aspect concerning the criterion. For instance, the sentence \textit{\textquotedblleft I will be back!"} expresses a positive polarity about a general criterion without specifying any aspect.

\paragraph{Compute the ITE values} 
We define a matrix \prev{$ITE$} for each expert to represent their natural language evaluations in a measurable way. Each element \prev{$ite^{k}_{ij}$} represents the opinion of the expert $e_{k}$ about the aspect category or criterion $c_{j}$ of the alternative $x_{i}$ expressed in the review $t^{k}_i$. We compute the $ite^{k}_{ij}$ values by combining the number of opinions expressed about the criterion $c_{j}$ ($\#total\_c_{j}$) considering the positive ($\#pos\_c_{j}$) and the negative ($\#neg\_c_{j}$) opinions. The individual textual evaluation values belong to the interval $[-\tau, \tau]$. Equation \ref{get_ite_kij} defines how to calculate each $ite^{k}_{ij}$.




\begin{equation}
ite^{k}_{ij} = \dfrac{\tau(\#pos\_c_{j} - \#neg\_c_{j})}{\#total\_c_{j}} 
\label{get_ite_kij}
\end{equation}



\new{We illustrate the calculation of the $ite^{k}_{ij}$ value with the following opinion given by the expert $e_{1}$ about the alternative $x_{1}$ as example: \textit{While sake and sushi were fantastic, rice was pasty}. The opinion speaks about two aspects about the criterion \textit{food} (sushi and rice), and one aspect from the criterion \textit{drinks} (sake). The DOC-ABSADeepL model returns three output vectors of the same size than the number of tokens of the input opinion, and they indicate whether a token is an aspect, the category of the aspect and the polarity of the opinion about that aspect. In this case, the DOC-ABSADeepL model returns three ten\footnote{The opinion is composed of ten tokens, nine words and one comma.} component output vectors, specifically the vector $\vec{aspects} = (0, 1, 0, 1, 0, 0, 0, 1, 0, 0)$, where the value 1 corresponds to the tokens: sake, sushi and rice; the vector $\vec{categories} = (0, 1, 0, 2, 0, 0, 0, 2, 0, 0)$, where $1$ refers to the criterion \textit{drinks} and $2$ refers to the criterion \textit{food}; and the vector $\vec{polarities} = (0, 1, 0, 1, 0, 0, 0, 2, 0 ,0)$, where $1$ means \textit{positive} and $2$ means \textit{negative}. Then, we set the value of $\tau$ to 2, and we calculate the $ite^{k}_{ij}$ value following the Equation \ref{get_ite_kij} as: \prev{$ite^{1}_{11} = \frac{2(1-0)}{1} = 2$} (\textit{drinks} category) and \prev{$ite^{1}_{12} = \frac{2(1-1)}{2} = 0$} (\textit{food} category).}

\subsubsection{Individual Numerical Evaluations  }\label{obtainINE}
An individual numerical evaluation matrix ($INE$) is defined for each expert \nuevo{only when the e-commerce site allows experts to provide numerical ratings. These matrices} collect the numerical ratings that experts may provide evaluating pre-defined criteria related to the alternatives. \nuevo{The alternatives are represented into the rows while the criteria are represented into the columns.}

\nuevo{Formally,} we define a matrix $INE$ for each expert, such that the element $ine^{k}_{ij}$ refers to the numerical evaluation to the alternative $x_{i}$ according to the criterion $c_{j}, j=1,...,m$ provided by the expert $e_k$. The $ine^{k}_{ij}$ values belong to \cris{the} interval $[-\tau, \tau]$.

\subsection{Alternative choice decisions}\label{selecting_best_phase}

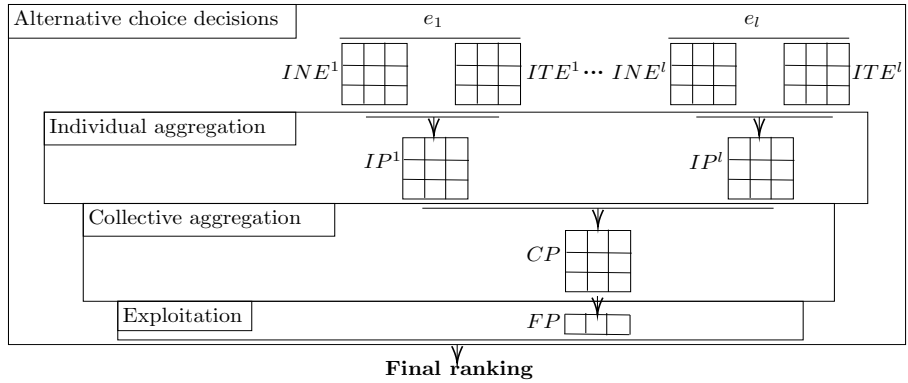
\begin{figure}[!h]
\tikzset{every picture/.style={line width=0.2pt}} 
\resizebox{\textwidth}{5cm}{      

\begin{tikzpicture}[x=0.75pt,y=0.75pt,yscale=-1,xscale=1]

\draw   (5,436) -- (180,436) -- (180,457.8) -- (5,457.8) -- cycle ;
\draw   (244.91,522.21) -- (284.53,522.21) -- (284.53,561.83) -- (244.91,561.83) -- cycle ;
\draw    (245.51,536.35) -- (285.17,536.86) ;
\draw    (245.08,549.09) -- (284.74,549.6) ;
\draw    (258.25,522.08) -- (258.25,561.15) ;
\draw    (270.99,521.65) -- (270.99,562) ;
\draw   (343.91,582.21) -- (383.53,582.21) -- (383.53,621.83) -- (343.91,621.83) -- cycle ;
\draw    (344.51,596.35) -- (384.17,596.86) ;
\draw    (344.08,609.09) -- (383.74,609.6) ;
\draw    (357.25,582.08) -- (357.25,621.15) ;
\draw    (369.99,581.65) -- (369.99,622) ;
\draw    (263.8,508.8) -- (263.8,518.8) ;
\draw [shift={(263.8,520.8)}, rotate = 270] [color={rgb, 255:red, 0; green, 0; blue, 0 }  ][line width=0.75]    (10.93,-3.29) .. controls (6.95,-1.4) and (3.31,-0.3) .. (0,0) .. controls (3.31,0.3) and (6.95,1.4) .. (10.93,3.29);
\draw    (222.5,508.8) -- (303.5,508.8) ;
\draw   (5,436) -- (550.8,436) -- (550.8,656) -- (5,656) -- cycle ;
\draw   (407.91,461.21) -- (447.53,461.21) -- (447.53,500.83) -- (407.91,500.83) -- cycle ;
\draw    (408.51,475.35) -- (448.17,475.86) ;
\draw    (408.08,488.09) -- (447.74,488.6) ;
\draw    (421.25,461.08) -- (421.25,500.15) ;
\draw    (433.99,460.65) -- (433.99,501) ;
\draw   (476.91,461.21) -- (516.53,461.21) -- (516.53,500.83) -- (476.91,500.83) -- cycle ;
\draw    (477.51,475.35) -- (517.17,475.86) ;
\draw    (477.08,488.09) -- (516.74,488.6) ;
\draw    (490.25,461.08) -- (490.25,500.15) ;
\draw    (502.99,460.65) -- (502.99,501) ;
\draw    (406.5,457.8) -- (516.5,457.8) ;
\draw    (256.8,567.8) -- (470.5,567.8) ;
\draw    (363.65,567.8) -- (363.65,577.8) ;
\draw [shift={(363.65,579.8)}, rotate = 270] [color={rgb, 255:red, 0; green, 0; blue, 0 }  ][line width=0.75]  (10.93,-3.29) .. controls (6.95,-1.4) and (3.31,-0.3) .. (0,0) .. controls (3.31,0.3) and (6.95,1.4) .. (10.93,3.29);
\draw    (343.51,636.35) -- (383.17,636.86) ;
\draw    (343.08,649.09) -- (382.74,649.6) ;
\draw    (356.25,635.6) -- (356.25,648.6) ;
\draw    (368.99,635.6) -- (368.99,650) ;
\draw    (383.17,636.86) -- (383.17,649.6) ;
\draw    (343.51,636.35) -- (343.51,649.09) ;
\draw    (278,656) -- (278,666) ;
\draw [shift={(278,668)}, rotate = 270] [color={rgb, 255:red, 0; green, 0; blue, 0 }  ][line width=0.75]    (10.93,-3.29) .. controls (6.95,-1.4) and (3.31,-0.3) .. (0,0) .. controls (3.31,0.3) and (6.95,1.4) .. (10.93,3.29)   ;

\draw   (207.91,461.21) -- (247.53,461.21) -- (247.53,500.83) -- (207.91,500.83) -- cycle ;
\draw    (208.51,475.35) -- (248.17,475.86) ;
\draw    (208.08,488.09) -- (247.74,488.6) ;
\draw    (221.25,461.08) -- (221.25,500.15) ;
\draw    (233.99,460.65) -- (233.99,501) ;
\draw   (276.91,461.21) -- (316.53,461.21) -- (316.53,500.83) -- (276.91,500.83) -- cycle ;
\draw    (277.51,475.35) -- (317.17,475.86) ;
\draw    (277.08,488.09) -- (316.74,488.6) ;
\draw    (290.25,461.08) -- (290.25,500.15) ;
\draw    (302.99,460.65) -- (302.99,501) ;
\draw    (206.5,457.8) -- (316.5,457.8) ;
\draw   (442.91,522.21) -- (482.53,522.21) -- (482.53,561.83) -- (442.91,561.83) -- cycle ;
\draw    (443.51,536.35) -- (483.17,536.86) ;
\draw    (443.08,549.09) -- (482.74,549.6) ;
\draw    (456.25,522.08) -- (456.25,561.15) ;
\draw    (468.99,521.65) -- (468.99,562) ;
\draw    (461.8,508.8) -- (461.8,518.8) ;
\draw [shift={(461.8,520.8)}, rotate = 270] [color={rgb, 255:red, 0; green, 0; blue, 0 }  ][line width=0.75]    (10.93,-3.29) .. controls (6.95,-1.4) and (3.31,-0.3) .. (0,0) .. controls (3.31,0.3) and (6.95,1.4) .. (10.93,3.29)   ;
\draw    (423.5,508.8) -- (506.5,508.8) ;
\draw   (26.79,505.67) -- (527.79,505.67) -- (527.79,564.78) -- (26.79,564.78) -- cycle ;
\draw   (50.5,564.71) -- (507.5,564.71) -- (507.5,628) -- (50.5,628) -- cycle ;
\draw   (71.5,628) -- (488.5,628) -- (488.5,653) -- (71.5,653) -- cycle ;
\draw   (71.5,628) -- (153.01,628) -- (153.01,647) -- (71.5,647) -- cycle ;
\draw    (363.34,624.6) -- (363.34,634.6) ;
\draw [shift={(363.34,636.6)}, rotate = 270] [color={rgb, 255:red, 0; green, 0; blue, 0 }  ][line width=0.75]    (10.93,-3.29) .. controls (6.95,-1.4) and (3.31,-0.3) .. (0,0) .. controls (3.31,0.3) and (6.95,1.4) .. (10.93,3.29)   ;
\draw   (26.79,505.67) -- (179.79,505.67) -- (179.79,526.86) -- (26.79,526.86) -- cycle ;
\draw   (50.5,564.71) -- (203.5,564.71) -- (203.5,585.9) -- (50.5,585.9) -- cycle ;

\draw (7,439) node [anchor=north west][inner sep=0.75pt]   [align=left] {Alternative choice decisions};
\draw (220,530) node [anchor=north west][inner sep=0.75pt]    {$IP^{1}$};
\draw (319,591.8) node [anchor=north west][inner sep=0.75pt]   [align=left] {$CP$};
\draw (371,472) node [anchor=north west][inner sep=0.75pt]    {$INE^{l}$};
\draw (517,472) node [anchor=north west][inner sep=0.75pt]    {$ITE^{l}$};
\draw (457.67,446.67) node   [align=left] {$e_{l}$};
\draw (360.91,480) node   [align=left] {\textbf{...}};
\draw (319,635) node [anchor=north west][inner sep=0.75pt]    {$FP$};
\draw (233,665) node [anchor=north west][inner sep=0.75pt]   [align=left] {\textbf{Final ranking}};
\draw (170,472) node [anchor=north west][inner sep=0.75pt]    {$INE^{1}$};
\draw (319,472) node [anchor=north west][inner sep=0.75pt]    {$ITE^{1}$};
\draw (263.67,446.67) node   [align=left] {$e_{1}$};
\draw (418,530) node [anchor=north west][inner sep=0.75pt]    {$IP^{l}$};
\draw (28.79,508.67) node [anchor=north west][inner sep=0.75pt]   [align=left] {Individual aggregation};
\draw (52.5,567.71) node [anchor=north west][inner sep=0.75pt]   [align=left] {Collective aggregation};
\draw (73.5,631) node [anchor=north west][inner sep=0.75pt]   [align=left] {Exploitation};

\end{tikzpicture}

}
\caption{The \nuevo{\textit{Alternative choice decisions}} phase of the \nuevo{SA-MpMcDM methodology.}}
\label{squeme_phase3}
\end{figure}

\czm{Once the expert knowledge is extracted and numerically represented, \nuevo{the alternative choice decision} is conducted \eugenio{through three tasks: \begin{enumerate*}[label=(\arabic*)]\item the individual aggregation of the \nuevo{$ITE$ and $INE$} matrices of each expert; \item the collective aggregation of the aggregated evaluations of the experts; and \item the exploitation of the collective aggregation for selecting the best alternative.\end{enumerate*} Figure \ref{squeme_phase3} shows the \nuevo{\textit{alternative choice decisions phase}}, and we subsequently detail it.}} 

\paragraph{\czm{Individual aggregation}} Each expert has associated two matrices that collect her evaluation. The \prev{$ITE^{k}$} matrix contains the \prev{individual} numerical evaluations inferred by the \nuevo{DOC-ABSADeepL model,} by extracting the expert knowledge from the \prev{textual} natural language evaluations that the expert freely provides. Optionally, the \prev{$INE^{k}$} matrix contains the \prev{individual} numerical evaluations that the expert provides directly on the \nuevo{pre-defined} criteria. We define \new{an Individual Preference ($IP$) matrix }
for each expert \geni{combining their corresponding  $ITE^{k}$ and $INE^{k}$ matrices.} \geni{Each $ip^{k}_{ij}$ value of the $IP^{k}$ matrix is computed according to Equation \ref{get_ip_kij}}
\begin{equation}
ip^{k}_{ij} = \omega_{ite} \cdot ite^{k}_{ij} + \omega_{ine} \cdot ine^{k}_{ij} 
\label{get_ip_kij}
\end{equation}
\nuevo{such} that \prev{$\omega_{ite}$ + $\omega_{ine}$ = 1}. The weight \prev{$\omega_{ite}$} represents the importance of the textual evaluation analysed by \nuevo{DOC-ABSADeepL,} while the weight \prev{$\omega_{ine}$} represents the importance of the numerical evaluation provided by the expert. 
By default, if the \nuevo{MpMc}DM problem does not specify \nuevo{the relevance of the textual and numerical evaluations}, we set \prev{$\omega_{ite}$ = $\omega_{ine}$ = 0.5}. When experts are not allowed to provide numerical evaluations, the weigth \prev{$\omega_{ine}$} takes value zero, so \prev{$ip^{k}_{ij}$} is just \prev{$ite^{k}_{ij}$}. In any case, the \prev{$ip^{k}_{ij}$} values belong to the interval $[-\tau, \tau]$.

\paragraph{\czm{Collective aggregation}} The particular evaluations of the experts \czm{are }
aggregated to obtain a collective evaluation for each alternative based on each criterion. The \new{Collective Preference ($CP$) matrix }
is defined for representing such collective evaluation. The element $cp_{ij}$ is defined by \prev{$\phi(ip^{k}_{ij})$}, where $\phi$ refers to the average operator that aggregate the evaluation of the $k$ experts. The $cp_{ij}$ values belong to the interval $[-\tau, \tau]$. 

\paragraph{\czm{Exploitation}} Once the collective evaluation of each alternative based on each criterion is obtained, \czm{we get the general assessment for each alternative. }
For this purpose, we aggregate the evaluation of the criteria for each alternative. The criteria have associated weights since not all of them have the same importance. Then, the set of criteria is linked to a set of weights $\omega=\{\omega_{1},…,\omega_{m}\}$. For example, users could evaluate restaurants according to the criteria \textit{food} and \textit{service}. However, \textit{food} is usually more important than \textit{service} when evaluating a restaurant, hence the weight associated to the \textit{food} criterion will be greater than the weight of \textit{service} criterion. 


The criteria weights \nuevo{could} be directly established based on the requirements of the problem. However, the \nuevo{SA-MpMcDM methodology} allows to establish them according to the reviews of the experts. \nuevo{We define the criteria weighting through the attention of the experts to set the criteria weights.} \nuevo{This procedure consists of counting} the number of evaluations, that is, the sum of the number of opinions and the number of numerical ratings of all the experts about each criterion, and \nuevo{dividing} each one by the total number of evaluations. This way, the criteria that receive 
more \nuevo{attention of the experts, that is they have more evaluations,} will have greater weight highlighting \cris{their} importance. Formally,
\begin{equation}
\omega_{j} = \dfrac{n\_evaluations(c_{j})}{n\_total}, j=1,...,m
\label{get_weigths}
\end{equation}
where $n\_evaluations(c_{j})$ is the sum of all the evaluations (opinions and numerical ratings) \new{about }
the criterion $c_{j}$, and $n\_total$ is the total number of evaluations.

For each alternative, the model aggregates the evaluation obtained for each criterion based on the \nuevo{criteria weighting (see Equation \ref{get_weigths})}. We define \new{the final preference (FP) vector }
representing the \prev{final preferences} for all the alternatives by means of the weighted average operator. The element \prev{$fp_{i}$} is computed by 
\begin{equation}
fp_{i} = \omega_{1} \cdot cp_{i1} + \dots +  \omega_{m} \cdot cp_{im}
\label{get_fp_i}
\end{equation}
where $\omega_{j} \in [0,1]$ and $\sum_{j=1}^{m}\omega_{j}=1$. Then, each alternative $x_{i}$ has an associated value\czm{, which belongs to $[-\tau,\tau]$,} collecting all the expert's evaluations. 
The ranking of the alternatives is obtained by ordering those values. The highest value represents the best alternative, while the lowest value represents the worst.

\subsection{Architecture of the methodology}\label{full_arq}
\nuevo{We outline the architecture of the SA-MpMcDM methodology in Figure \ref{squeme_phases}.}

\input{tikzs/squeme_Phases.tex}
    
\section{Case of study: choice of a restaurant}\label{case_study}
We consider the selection of a restaurant taking into account the reviews of several restaurants in an e-commerce site as \nuevo{an MpMcDM} problem. Subsequently, we detail the real use case, and how we use the \nuevo{SA-MpMcDM methodology} for ranking restaurants located in the city of London.



Restaurants are evaluated by experts through TripAdvisor. This is a well-known e-commerce site\cris{,} which allows both making reservations and providing travel related reviews, \textit{e.g.}, reviews evaluating restaurants. \geni{Also,} experts can provide their reviews both by \czm{written texts} and by numerical evaluations. The numerical evaluations are discrete variables that can take five values, that is, in a scale of five levels of opinion intensity. Then, $\tau=2$ according to our \nuevo{methodology}. \nuevo{SA-MpMcDM} allows to analyze all possible sources of information, \textit{i.e.}, the written texts and the numerical ratings.

\eugenio{The restaurant reviews in TripAdvisor may be written in any language spoken in the world, but we only processed the reviews written in English in this evaluation.} The set of criteria is defined by \prev{$C=\{$\textit{Restaurant}, \textit{Food}, \textit{Service}, \textit{Drinks}, \textit{Ambience}, \textit{Location}$\}$} \geni{because}:
\begin{enumerate*}[label=(\arabic*)]
    \item TripAdvisor considers \textit{Food} and \textit{Service} as numerical evaluation criteria;
    \item \textit{Drinks}, \textit{Ambience} and \textit{Location} are criteria usually mentioned by users when evaluating restaurants; and \item \textit{Restaurant} criteri\cris{on}, which is a general criteri\cris{on} that gathers the aspects of a restaurant. Furthermore, TripAdvisor allows to provide an overall numerical evaluation to the restaurants, which is represented in the \textit{Restaurant} criteri\cris{on}.
\end{enumerate*}





The exposed problem is solved through the \nuevo{SA-MpMcDM methodology} as we describe in the subsequent sections.

\subsection{Obtaining input evaluations: TripR-2020 dataset}\label{input_data_exp}


\czm{To solve the \nuevo{MpMcDM} case study, we download the restaurant reviews from TripAdvisor using a web crawler. The developed web crawler considers as entities restaurants located in \cz{London}. Given the link of a particular restaurant, the web crawler download\cris{s} all its reviews providing attributes concerning the restaurant, the user and the review such as the \textit{body}, the \textit{title} and the numerical ratings. 
We download the reviews of 78 restaurants from the \textit{Local Cuisine} section. The 78 restaurants are evaluated by 1,428 experts. Unfortunately, most of them evaluate just one or two restaurants. Therefore, we select the experts who have evaluated a number of restaurants in common. For this case study, we limit such number to four. Of the 1,428 original experts, only six experts evaluate four restaurants in common. There is not loss of information since the 6 experts evaluate the 4 restaurants gathering a dataset with 24 reviews.

We build \nuevo{and release} the TripR-2020 dataset\footnote{\nuevo{The TripR-2020 dataset provides the evaluations of the 1,428 experts. Moreover, it provides the 24 evaluations of the 6 experts considered in this case of study. We provide those 24 evaluations as raw text and annotated opinions for processing the evaluation of the experts. We describe the annotation of the evaluations in Section \ref{data_preparation_exp}.}} collecting the numerical evaluations and the written texts downloaded by the web crawler. Written texts are composed by a \textit{title} and a \textit{body}. 
Then, the TripR-2020 provides the data of the case study which could be use to evaluate \nuevo{the SA-MpMcDM methodology.}
}

Table \ref{set_exp_alt} shows \nuevo{the experts and the alternatives from the case of study. } 
The set of alternatives is given by the restaurants $X = \{x_{1}, x_{2}, x_{3}, x_{4}\}$ = $\{${\textit{Oxo Tower Restaurant, Bar and Brasserie\footnote{\url{https://www.tripadvisor.co.uk/Restaurant_Review-g186338-d680501}}}},
{\textit{J. Sheekey\footnote{\url{https://www.tripadvisor.co.uk/Restaurant_Review-g186338-d719379}}}},
{\textit{The Wolseley\footnote{\url{https://www.tripadvisor.co.uk/Restaurant_Review-g186338-d731402}}}},
{\textit{The Ivy\footnote{\url{https://www.tripadvisor.co.uk/Restaurant_Review-g186338-d776287}}}}$\}$. The set of experts is represented by $E=\{e_{1}, e_{2}, e_{3}, e_{4}, e_{5}, e_{6}\}$.\footnote{For privacy reasons, we do not publish the names and identifiers of the experts who evaluate the restaurants.} 

\begin{table}[!h]
\centering
\small
\begin{tabularx}{\textwidth}{@{\hspace{2em}}ccc@{\hspace{5em}}cc}
\toprule
	& \textbf{Experts} &  & \multicolumn{2}{c}{\textbf{Alternatives}}\\
\midrule
e\textsubscript{1}      & e\textsubscript{2}      & e\textsubscript{3}      & x\textsubscript{1} = Oxo Tower Restaurant & x\textsubscript{2} = J. Sheekey\\
e\textsubscript{4}      & e\textsubscript{5}      & e\textsubscript{6}      & x\textsubscript{3} = The Wolseley         &  x\textsubscript{4} = The Ivy  \\
\bottomrule
\end{tabularx}
\caption{\protect\label{set_exp_alt}Set of experts and set of alternatives of the case of study \nuevo{collected into the TripR-2020 dataset. It} has 24 reviews since the six experts evaluate the four restaurants.\protect}
\end{table}

\begin{figure}[!h]
\centering
\tikzset{every picture/.style={line width=0.5pt}} 
\resizebox{\textwidth}{5.5cm}{     
\begin{tikzpicture}[x=0.75pt,y=0.75pt,yscale=-1,xscale=1]

\draw   (193.59,18.25) .. controls (193.59,9.72) and (200.51,2.8) .. (209.04,2.8) .. controls (217.58,2.8) and (224.5,9.72) .. (224.5,18.25) .. controls (224.5,26.79) and (217.58,33.71) .. (209.04,33.71) .. controls (200.51,33.71) and (193.59,26.79) .. (193.59,18.25) -- cycle ; \draw   (202.24,13) .. controls (202.24,12.15) and (202.93,11.45) .. (203.79,11.45) .. controls (204.64,11.45) and (205.33,12.15) .. (205.33,13) .. controls (205.33,13.85) and (204.64,14.54) .. (203.79,14.54) .. controls (202.93,14.54) and (202.24,13.85) .. (202.24,13) -- cycle ; \draw   (212.75,13) .. controls (212.75,12.15) and (213.44,11.45) .. (214.3,11.45) .. controls (215.15,11.45) and (215.84,12.15) .. (215.84,13) .. controls (215.84,13.85) and (215.15,14.54) .. (214.3,14.54) .. controls (213.44,14.54) and (212.75,13.85) .. (212.75,13) -- cycle ; \draw   (201.31,24.44) .. controls (206.47,28.56) and (211.62,28.56) .. (216.77,24.44) ;
\draw    (104.95,65.71) -- (186,66.47) ;
\draw    (104.08,84.65) -- (185.13,85.41) ;
\draw    (130.99,66) -- (130.99,83.92) ;
\draw    (157.02,66) -- (157.02,86) ;
\draw    (186,66.47) -- (186,85.41) ;
\draw    (104.95,65.71) -- (104.95,84.65) ;
\draw    (222,35) -- (233.61,46.61) ;
\draw [shift={(235.02,48.02)}, rotate = 225] [color={rgb, 255:red, 0; green, 0; blue, 0 }  ][line width=0.75]    (10.93,-3.29) .. controls (6.95,-1.4) and (3.31,-0.3) .. (0,0) .. controls (3.31,0.3) and (6.95,1.4) .. (10.93,3.29)   ;
\draw    (193.02,34.02) -- (180.44,46.61) ;
\draw [shift={(179.02,48.02)}, rotate = 315] [color={rgb, 255:red, 0; green, 0; blue, 0 }  ][line width=0.75]    (10.93,-3.29) .. controls (6.95,-1.4) and (3.31,-0.3) .. (0,0) .. controls (3.31,0.3) and (6.95,1.4) .. (10.93,3.29)   ;
\draw    (15.41,277.19) -- (46.46,277.59) ;
\draw    (15.08,287.16) -- (46.12,287.56) ;
\draw    (25.39,276.6) -- (25.39,286.78) ;
\draw    (35.36,276.6) -- (35.36,287.87) ;
\draw    (46.46,277.59) -- (46.46,287.56) ;
\draw    (15.41,277.19) -- (15.41,287.16) ;
\draw    (54,281) -- (62.81,281) ;
\draw    (54,286.03) -- (62.81,286.03) ;
\draw    (54,283.52) -- (62.81,283.52) ;
\draw   (206,54.19) -- (230.29,54.19) -- (230.29,77) -- (206,77) -- cycle ;
\draw    (103.95,114.71) -- (185,115.47) ;
\draw    (103.08,133.65) -- (184.13,134.41) ;
\draw    (129.99,115) -- (129.99,132.92) ;
\draw    (156.02,115) -- (156.02,135) ;
\draw    (185,115.47) -- (185,134.41) ;
\draw    (103.95,114.71) -- (103.95,133.65) ;
\draw   (206,107.19) -- (230.29,107.19) -- (230.29,130) -- (206,130) -- cycle ;
\draw    (104.95,164.71) -- (186,165.47) ;
\draw    (104.08,183.65) -- (185.13,184.41) ;
\draw    (130.99,165) -- (130.99,182.92) ;
\draw    (157.02,165) -- (157.02,185) ;
\draw    (186,165.47) -- (186,184.41) ;
\draw    (104.95,164.71) -- (104.95,183.65) ;
\draw   (205,159.19) -- (229.29,159.19) -- (229.29,182) -- (205,182) -- cycle ;
\draw   (205,211.19) -- (629,211.19) -- (629,251.46) .. controls (364,251.46) and (417,265.98) .. (205,256.58) -- cycle ;
\draw    (105.95,212.71) -- (187,213.47) ;
\draw    (105.08,231.65) -- (186.13,232.41) ;
\draw    (131.99,213) -- (131.99,230.92) ;
\draw    (158.02,213) -- (158.02,233) ;
\draw    (187,213.47) -- (187,232.41) ;
\draw    (105.95,212.71) -- (105.95,231.65) ;
\draw   (205,211.19) -- (229.29,211.19) -- (229.29,234) -- (205,234) -- cycle ;
\draw   (205,159.19) -- (629,159.19) -- (629,199.46) .. controls (356.96,199.46) and (407.61,213.98) .. (205,204.58) -- cycle ;
\draw   (206,107.19) -- (629,107.19) -- (629,147.46) .. controls (357.96,147.46) and (408.61,161.98) .. (206,152.58) -- cycle ;
\draw   (206,54.19) -- (629,54.19) -- (629,94.46) .. controls (357.96,94.46) and (408.61,108.98) .. (206,99.58) -- cycle ;

\draw (42,40) node   [align=left] {{\footnotesize restaurant}};
\draw (39.5,76.55) node   [align=left] {{\small Oxo Tower}};
\draw (217,65.59) node   [align=left] {$t^{1}_{1}$};
\draw (67,278) node [anchor=north west][inner sep=0.75pt]   [align=left] {{\footnotesize numerical ratings ($\in [-2,2]$) evaluating the criteria \textit{restaurant}, \textit{food} and \textit{service}}};
\draw (180.67,17.67) node   [align=left] {e{\scriptsize 1}};
\draw (6,120) node [anchor=north west][inner sep=0.75pt]   [align=left] {J. Sheekey};
\draw (5,167) node [anchor=north west][inner sep=0.75pt]   [align=left] {The Wolseley};
\draw (6,215) node [anchor=north west][inner sep=0.75pt]   [align=left] {The Ivy};
\draw (208,61) node [anchor=north west][inner sep=0.75pt]   [align=left] {{\scriptsize \textbf{ \ \ \ \ \ Table with a view. }If the weather is good, make sure you ask for a table outside. }\\[-0.5em]{\scriptsize We ate in the summer when the sun was setting over London. Really was amazing.}};
\draw (207,167) node [anchor=north west][inner sep=0.75pt]   [align=left] {{\scriptsize  \ \ \ \ \ \ \ \ \textbf{Perfect for Breakfast.} Love The Wolseley for its location in }\\[-0.5em]{\scriptsize Central London with a great breakfast. Staff always friendly and helpful.}};
\draw (112,72) node [anchor=north west][inner sep=0.75pt]   [align=left] {{\scriptsize 1} \ \ \ {\scriptsize  NA \ \ \ NA}};
\draw (207,219) node [anchor=north west][inner sep=0.75pt]   [align=left] {{\scriptsize  \ \ \ \ \ \ \ \textbf{Consistently Good.} I have been coming for years. Always good}\\[-0.5em]{\scriptsize atmosphere and fun people watching. The food is always good and quick.}};
\draw (217,118.59) node   [align=left] {$t^{1}_{2}$};
\draw (111,120) node [anchor=north west][inner sep=0.75pt]   [align=left] {{\scriptsize 2 } \ \ {\scriptsize  NA \ \ \ NA}};
\draw (217,170.59) node   [align=left] {$t^{1}_{3}$};
\draw (112,171) node [anchor=north west][inner sep=0.75pt]   [align=left] {{\footnotesize 1} \ \ \ {\scriptsize  NA \ \ \ NA}};
\draw (208,115) node [anchor=north west][inner sep=0.75pt]   [align=left] {{\scriptsize \textbf{ \ \ \ \ \ \ Sit at the bar.} Such a great restaurant. Love Marco (one of the managers) }\\[-0.5em] {\scriptsize he used to work at E\&O. Perfect place for fish - always try and sit and eat at the bar.}};
\draw (217,222.59) node   [align=left] {$t^{1}_{4}$};
\draw (113,219) node [anchor=north west][inner sep=0.75pt]   [align=left] {{\footnotesize 1} \ \ \ {\scriptsize  NA \ \ \ NA}};

\end{tikzpicture}

}
\caption{Original input evaluations provided by the expert $e_{1}$ from the TripR-2020 dataset. 
The expert does not provide numerical ratings for \textit{food} and \textit{service} criteria.
}
\label{example_e1}
\end{figure}
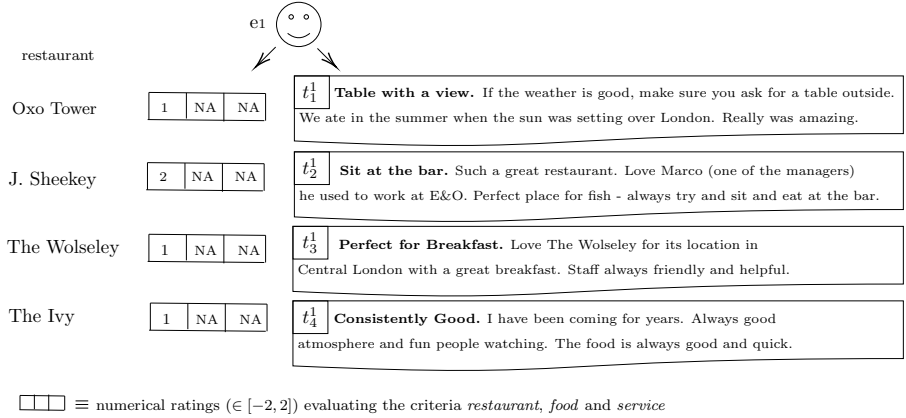

Figure \ref{example_e1} shows the information related to the expert $e_{1}$ downloaded by the web crawler and collected into the TripR-2020 dataset. The expert provides numerical ratings \nuevo{just} on the general \textit{restaurant} criteri\cris{on}. However, the expert provides so much more information into the written texts for all the restaurants.

\subsection{Distilling opinions at criterion level: DOC-ABSADeepL model}\label{data_preparation_exp}

\begin{figure}[!h]
\centering
\tikzset{every picture/.style={line width=0.1pt}} 
\resizebox{0.7\textwidth}{2cm}{  

\begin{tikzpicture}[x=0.75pt,y=0.75pt,yscale=-1,xscale=1]

\draw    (88,21) -- (88,91.33) ;
\draw    (88,21) -- (97.5,21) ;
\draw    (88,91.33) -- (97.5,91.33) ;
\draw    (382.5,21) -- (392,21) ;
\draw    (382.5,91.33) -- (392,91.33) ;
\draw    (392,21) -- (392,91.33) ;

\draw (89,4) node [anchor=north west][inner sep=0.75pt]   [align=left] {{\small restaurant \ \ food \ \ service \ \ drinks \ \ ambience \ \ location}};
\draw (4,23) node [anchor=north west][inner sep=0.75pt]   [align=left] {{\small Oxo Tower}};
\draw (4,43) node [anchor=north west][inner sep=0.75pt]   [align=left] {{\small J. Sheekey}};
\draw (4,63) node [anchor=north west][inner sep=0.75pt]   [align=left] {{\small The Wolseley}};
\draw (4,83) node [anchor=north west][inner sep=0.75pt]   [align=left] {{\small The Ivy}};
\draw (99,37) node [anchor=north west][inner sep=0.75pt]   [align=left] {{\large ITE, INE, IP and CP matrices values}\\{\large belongs to the interval $[-\tau, \tau]$ = $[-2,2]$}};
\draw (397,89) node [anchor=north west][inner sep=0.75pt]   [align=left] {{\tiny 4x6}};

\end{tikzpicture}
}
\caption{Format of all the matrices obtained by \nuevo{SA-MpMcDM }for the case of study.}
\label{example_matrices}
\end{figure}
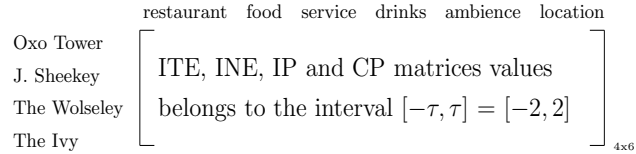

\nuevo{This section analyzes the written texts from the TripR-2020 dataset and collects the numerical evaluations generating the $ITE$ and the $INE$ matrices, respectively.} Figure \ref{example_matrices} shows the format of all the matrices generated for this case study. The rows represent the expert evaluations provided to the restaurants \textit{Oxo Tower}, \textit{J. Sheekey}, \textit{The Wolseley} and \textit{The Ivy} respectively. The columns represent the evaluation for all the criteria, \textit{i.e.},  \textit{restaurant},  \textit{food}, \prev{ \textit{service}}, \textit{drinks},  \textit{ambience} and  \textit{location} respectively.

\nuevo{The $ITE$ matrices are built extracting the opinions using the DOC-ABSADeepL model, and transforming them into numbers as described in Section \ref{sentiment_analysis_phase}. Then, we first describe how to extract the opinions from the written texts, and then how to compute the $ite$ values.}


\paragraph{Extract the opinions} To extract the opinions from the vectors $T^{k}$, the \nuevo{DOC-ABSADeepL model} is fed with such vectors. To represent each input word by its word embedding
, we consider the Fasttext word embedddings \cite{joulin-etal-2017-bag} trained on Common Crawl \cite{crawl-corpus}. The word embedding dimension is $d = 300$.\footnote{\url{https://fasttext.cc/docs/en/english-vectors.html}} \nuevo{DOC-ABSADeepL} is trained on \geni{the restaurant training set of the ABSA task of} SemEval-2016.\footnote{\url{http://alt.qcri.org/semeval2016/task5/}} This dataset contains multiple restaurant reviews that are splitted into sentences. Most of the sentences in this dataset do not exceed 200 words, so we set $s = 200$ words as input for the multi-task deep learning model. Remaining hyper-parameters from the end-to-end deep learning model are $h_{lstm}=128$, $k_{1}=2$ and $fm_{1}=100$. To sum up, Table \ref{t_hyperparameters} presents the hyper-parameters established for \nuevo{DOC-ABSADeepL}.


\begin{table}[!h]
\centering
\small
\begin{tabular*}{\textwidth}{@{\extracolsep{\fill}}lr@{}}
\toprule
\textbf{Hyperparameter}	&	\textbf{Value}\\
\midrule
Input sentence size ($s$)		&	200\\
Word embedding dimension ($d$)	&	300\\
LSTM hidden units ($h_{lstm}$)	&	128\\
CNN kernel size ($k_{1}$)	&	2\\
CNN feature maps ($fm_{1}$)	&	100\\
batch size 	&	16\\
epochs 	&	20\\
\bottomrule
\end{tabular*}
\caption{Hyperparameters of the \nuevo{DOC-ABSADeepL} model.}
\label{t_hyperparameters}
\end{table}





To extract the expert knowledge from the written texts of the TripR-2020 dataset, the $T^{k}$ vectors are transformed into a suitable format for the \nuevo{DOC-ABSADeepL model}. This format is analogous to the training set of SemEval-2016\nuevo{, because we use this dataset for training DOC-ABSADeepL}. Therefore, \begin{enumerate*}[label=(\arabic*)] 
\item we split the written texts by sentences and 
\item we carry out a manual \eugenio{annotation} of the 24 reviews \geni{following} the official SemEval-2016 annotation manual. 
\end{enumerate*} \nuevo{Table \ref{exampleAnnotation} shows the manual annotation labeling of the \prev{$t^{1}_{4}$} natural language review.}
\begin{table}[!h]
\centering
\small
\begin{tabularx}{\textwidth}{@{}X@{\hspace{1em}}l@{\hspace{1.2em}}l@{\hspace{1.2em}}l@{}}
\toprule
\textbf{Sentence}	&	\textbf{aspect}	&	\textbf{category} & \textbf{polarity}\\
\midrule
\scriptsize Consistently good. & \footnotesize implicit &  \footnotesize restaurant &  \footnotesize positive \\
\midrule
\scriptsize I have been coming for years.	&  \footnotesize implicit &  \footnotesize restaurant &  \footnotesize positive\\
\midrule
\scriptsize Always good atmosphere and fun people watching. &  \footnotesize atmosphere &  \footnotesize ambience &  \footnotesize positive \\
\midrule
\scriptsize The food is always good and quick. &  \footnotesize food &  \footnotesize food &  \footnotesize positive 
\\
\bottomrule
\end{tabularx}
\caption{Annotation at aspect and opinion level of the review $t^{1}_{4}$. The review is split into sentences.}
\label{exampleAnnotation}
\end{table}

\geni{We first evaluate \nuevo{the DOC-ABSADeepL} model using the training and test subsets of the SemEval-2016 dataset. We use the F1-score and the Accuracy evaluation measures for the aspect-term extraction and polarity classification tasks respectively, as in SemEval-2016. The task of aspect category classification is not considered in SemEval-2016, so we use the F1-score as for the aspect term extraction task.\footnote{For the sake of clarity, the task 5 of SemEval-2016 defines aspect categories in a different way from us. We consider aspect categories as a cluster of aspect terms, while SemEval-2016 defines them as the combination of a cluster of aspect terms and the qualitative attribute that receives the opinion. For instance, we only consider the aspect category \textit{food}, and SemEval-2016 takes into account \textit{food\#quality}, \textit{food\#style} among others.} Although our aim is only to assess the \nuevo{SA-MpMcDM} methodology, we stress out that we reached the highest result in the aspect term extraction task according to the SemEval-2016 results \cite{pontiki2016semeval}.}

\geni{Once we evaluated the \nuevo{DOC-ABSADeepL} model on the SemEval-2016 dataset, we evaluated it using the TripR-2020 dataset as test set. Table \ref{resultsNetwork} shows the results reached by \nuevo{DOC-ABSADeepL} on both datasets. We highlight that the results reached on the TripR-2020 dataset are higher than using the test set of SemEval-2016, which means that \nuevo{the DOC-ABSADeepL} model is able to generalize the knowledge learnt from the training data.}

\begin{table}[!h]
\centering
\small
\begin{tabularx}{\textwidth}{@{}X@{\hspace{1em}}c@{\hspace{6em}}c@{}}
\toprule
\textbf{}	&	\textbf{SemEval-2016}	&	\textbf{\czm{TripR-2020}}\\
\midrule
Aspects \scriptsize{(\%F1)} & 72.53 
& 70.08 \\
\midrule
Categories \scriptsize{(\%F1)}	& 68.74 
& 78.31\\
\midrule
Polarities \scriptsize{(\%accuracy)} &	72.03	
& 80.12 
\\
\bottomrule
\end{tabularx}
\caption{Results of the \nuevo{DOC-ABSADeepL model} trained and tested in SemEval-2016 and tested in TripR-2020. \cz{Aspects and categories are evaluated according to \%F1-score while polarities are evaluated according to \%accuracy.}}
\label{resultsNetwork}
\end{table}


\paragraph{\czm{Compute the ITE values}}
Applying the Equation \ref{get_ite_kij}, we obtain the 
$ITE^{k}$ \czm{matrices shown in Figure \ref{cs_ite}} with the inferred numerical evaluations for each criterion. \czm{Figure \ref{example_matrices} shows the format of the matrices.} Analyzing as example the expert $e_{1}$ based on the restaurant $x_{4}$, we conclude that this expert provides very positive opinions about the criteria \textit{restaurant}, \textit{food} and \textit{ambience} when evaluating through written texts. Therefore, the expert provides more information through written texts than by providing numerical ratings.

\begin{figure}[!h]
\tiny \[
ITE^{1} =
    \begin{bmatrix}
    2 & NA & NA & NA & NA & 2 \\
    2 & 2 & NA & NA & NA & NA \\
    2 & NA & 2 & NA & NA & 2 \\
    2 & 2 & NA & NA & 2 & NA 
    \end{bmatrix}
\hspace{0.8cm}
ITE^{2} =
    \begin{bmatrix}
    2 & 2 & 2 & NA & 2 & 2 \\
    2 & 2 & NA & NA & NA & NA \\
    2 & 2 & 2 & NA & 2 & NA \\
    2 & 2 & 2 & NA & NA & NA 
    \end{bmatrix}
\]

\[
ITE^{3} =
    \begin{bmatrix}
    2 & 1.5 & NA & 2 & 2 & NA \\
    2 & 2 & 2 & NA & 2 & NA \\
    2 & 1.56 & NA & 2 & NA & NA \\
    2 & 2 & 2 & NA & 2 & NA 
    \end{bmatrix}
\hspace{0.3cm}
ITE^{4} =
    \begin{bmatrix}
    2 & 2 & 2 & 2 & 2 & 2 \\
    2 & 2 & NA & NA & NA & NA \\
    2 & NA & 2 & 2 & 2 & 2 \\
    0.4 & 1.5 & 0 & NA & NA & NA 
    \end{bmatrix}
\]

\[
ITE^{5} =
    \begin{bmatrix}
    2 & 2 & NA & NA & 2 & 2 \\
    2 & 2 & 2 & NA & NA & 2 \\
    2 & 2 & 2 & NA & 2 & NA \\
    2 & 2 & NA & NA & 2 & NA 
    \end{bmatrix}
\hspace{0.9cm}
ITE^{6} =
    \begin{bmatrix}
    2 & 2 & 2 & NA & 2 & NA \\
    2 & 2 & -2 & NA & 2 & NA \\
    2 & 2 & 2 & NA & 2 & NA \\
    2 & NA & 2 & NA & NA & NA 
    \end{bmatrix}
    \]
\normalsize

\caption{Individual Textual Evaluations matrices obtained by \nuevo{SA-MpMcDM}.\protect\label{cs_ite}}
\end{figure}

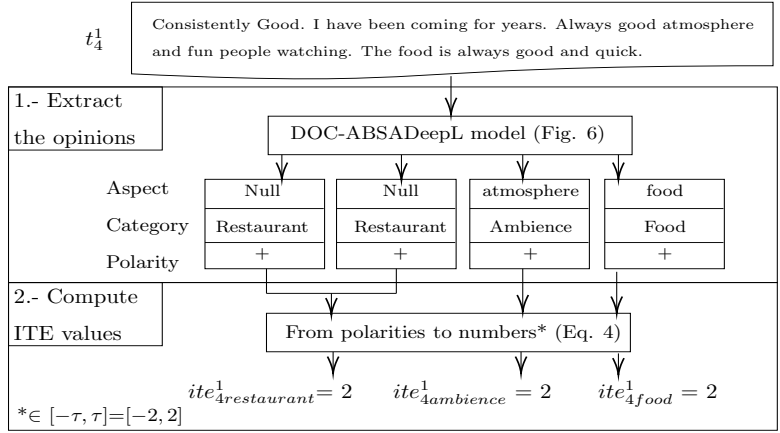
\begin{figure}[!h]
\centering
\tikzset{every picture/.style={line width=0.2pt}} 
\resizebox{0.86\textwidth}{5.7cm}{    
\begin{tikzpicture}[x=0.75pt,y=0.75pt,yscale=-1,xscale=1]

\draw   (78,5) -- (453.5,5) -- (453.5,46.25) .. controls (218.81,46.25) and (265.75,61.13) .. (78,51.5) -- cycle ;
\draw   (6,60) -- (456,60) -- (456,187.07) -- (6,187.07) -- cycle ;
\draw    (265,53) -- (265,76) ;
\draw [shift={(265,78)}, rotate = 270] [color={rgb, 255:red, 0; green, 0; blue, 0 }  ][line width=0.75]    (10.93,-3.29) .. controls (6.95,-1.4) and (3.31,-0.3) .. (0,0) .. controls (3.31,0.3) and (6.95,1.4) .. (10.93,3.29)   ;
\draw   (158,79) -- (371,79) -- (371,104) -- (158,104) -- cycle ;
\draw   (6,60) -- (96,60) -- (96,101) -- (6,101) -- cycle ;
\draw   (121,120) -- (191,120) -- (191,178.07) -- (121,178.07) -- cycle ;
\draw    (121,139) -- (190.5,139) ;
\draw    (121,161.07) -- (190.5,161.07) ;
\draw   (198,120) -- (268,120) -- (268,178.07) -- (198,178.07) -- cycle ;
\draw    (198,139) -- (267.5,139) ;
\draw    (198,161.07) -- (267.5,161.07) ;
\draw   (276,120) -- (346,120) -- (346,178.07) -- (276,178.07) -- cycle ;
\draw    (276,139) -- (345.5,139) ;
\draw    (276,161.07) -- (345.5,161.07) ;
\draw   (355,120) -- (425,120) -- (425,178.07) -- (355,178.07) -- cycle ;
\draw    (355,139) -- (424.5,139) ;
\draw    (355,161.07) -- (424.5,161.07) ;
\draw    (166,104.07) -- (166,114.07) -- (166,118.07) ;
\draw [shift={(166,120.07)}, rotate = 270] [color={rgb, 255:red, 0; green, 0; blue, 0 }  ][line width=0.75]    (10.93,-3.29) .. controls (6.95,-1.4) and (3.31,-0.3) .. (0,0) .. controls (3.31,0.3) and (6.95,1.4) .. (10.93,3.29)   ;
\draw    (236,104.07) -- (236,118.53) ;
\draw [shift={(236,120.53)}, rotate = 270] [color={rgb, 255:red, 0; green, 0; blue, 0 }  ][line width=0.75]    (10.93,-3.29) .. controls (6.95,-1.4) and (3.31,-0.3) .. (0,0) .. controls (3.31,0.3) and (6.95,1.4) .. (10.93,3.29)   ;
\draw    (307,104.07) -- (307,117.07) ;
\draw [shift={(307,119.07)}, rotate = 270] [color={rgb, 255:red, 0; green, 0; blue, 0 }  ][line width=0.75]    (10.93,-3.29) .. controls (6.95,-1.4) and (3.31,-0.3) .. (0,0) .. controls (3.31,0.3) and (6.95,1.4) .. (10.93,3.29)   ;
\draw    (363,104.07) -- (363,118.07) ;
\draw [shift={(363,120.07)}, rotate = 270] [color={rgb, 255:red, 0; green, 0; blue, 0 }  ][line width=0.75]    (10.93,-3.29) .. controls (6.95,-1.4) and (3.31,-0.3) .. (0,0) .. controls (3.31,0.3) and (6.95,1.4) .. (10.93,3.29)   ;
\draw    (157,178.07) -- (157,194.07) ;
\draw    (233,178.07) -- (233,194.07) ;
\draw    (157,194.07) -- (233,194.07) ;
\draw    (195,194.07) -- (195,206.07) ;
\draw [shift={(195,208.07)}, rotate = 270] [color={rgb, 255:red, 0; green, 0; blue, 0 }  ][line width=0.75]    (10.93,-3.29) .. controls (6.95,-1.4) and (3.31,-0.3) .. (0,0) .. controls (3.31,0.3) and (6.95,1.4) .. (10.93,3.29)   ;
\draw    (307,178.07) -- (307,204.07) ;
\draw [shift={(307,206.07)}, rotate = 270] [color={rgb, 255:red, 0; green, 0; blue, 0 }  ][line width=0.75]    (10.93,-3.29) .. controls (6.95,-1.4) and (3.31,-0.3) .. (0,0) .. controls (3.31,0.3) and (6.95,1.4) .. (10.93,3.29)   ;
\draw    (362,180.07) -- (362,205.07) ;
\draw [shift={(362,207.07)}, rotate = 270] [color={rgb, 255:red, 0; green, 0; blue, 0 }  ][line width=0.75]    (10.93,-3.29) .. controls (6.95,-1.4) and (3.31,-0.3) .. (0,0) .. controls (3.31,0.3) and (6.95,1.4) .. (10.93,3.29)   ;
\draw   (6,187.07) -- (456,187.07) -- (456,283) -- (6,283) -- cycle ;
\draw   (6,187.07) -- (96,187.07) -- (96,228.07) -- (6,228.07) -- cycle ;
\draw   (157,207) -- (370,207) -- (370,232) -- (157,232) -- cycle ;
\draw    (196,232.07) -- (196,244.07) ;
\draw [shift={(196,246.07)}, rotate = 270] [color={rgb, 255:red, 0; green, 0; blue, 0 }  ][line width=0.75]    (10.93,-3.29) .. controls (6.95,-1.4) and (3.31,-0.3) .. (0,0) .. controls (3.31,0.3) and (6.95,1.4) .. (10.93,3.29)   ;
\draw    (306,232.07) -- (306,244.07) ;
\draw [shift={(306,246.07)}, rotate = 270] [color={rgb, 255:red, 0; green, 0; blue, 0 }  ][line width=0.75]    (10.93,-3.29) .. controls (6.95,-1.4) and (3.31,-0.3) .. (0,0) .. controls (3.31,0.3) and (6.95,1.4) .. (10.93,3.29)   ;
\draw    (363,233.07) -- (363,245.07) ;
\draw [shift={(363,247.07)}, rotate = 270] [color={rgb, 255:red, 0; green, 0; blue, 0 }  ][line width=0.75]    (10.93,-3.29) .. controls (6.95,-1.4) and (3.31,-0.3) .. (0,0) .. controls (3.31,0.3) and (6.95,1.4) .. (10.93,3.29)   ;

\draw (89,15) node [anchor=north west][inner sep=0.75pt]   [align=left] {{\scriptsize Consistently Good. I have been coming for years. Always good atmosphere }\\[-0.5em]{\scriptsize and fun people watching. The food is always good and quick.}};
\draw (170,84) node [anchor=north west][inner sep=0.75pt]   [align=left] {{\small DOC-ABSADeepL model (Fig. 6)}};
\draw (125,122) node [anchor=north west][inner sep=0.75pt]   [align=left] {\begin{minipage}[lt]{42.64164400000001pt}\setlength\topsep{0pt}
\begin{center}
{\footnotesize Null}\\{\footnotesize Restaurant}\\\vspace{-0.2cm}{\footnotesize +}
\end{center}

\end{minipage}};
\draw (62,120) node [anchor=north west][inner sep=0.75pt]   [align=left] {{\small Aspect}\\{\small Category}\\{\small Polarity}};
\draw (8,63) node [anchor=north west][inner sep=0.75pt]   [align=left] {1.- Extract \\the opinions};
\draw (206,122) node [anchor=north west][inner sep=0.75pt]   [align=left] {\begin{minipage}[lt]{42.64164400000001pt}\setlength\topsep{0pt}
\begin{center}
{\footnotesize Null}\\{\footnotesize Restaurant}\\\vspace{-0.2cm}{\footnotesize +}
\end{center}

\end{minipage}};
\draw (280,122) node [anchor=north west][inner sep=0.75pt]   [align=left] {\begin{minipage}[lt]{45.815000000000005pt}\setlength\topsep{0pt}
\begin{center}
{\footnotesize atmosphere}\\{\footnotesize Ambience}\\\vspace{-0.2cm}{\footnotesize +}
\end{center}

\end{minipage}};
\draw (375,122) node [anchor=north west][inner sep=0.75pt]   [align=left] {\begin{minipage}[lt]{21.320856pt}\setlength\topsep{0pt}
\begin{center}
{\footnotesize food}\\{\footnotesize Food}\\\vspace{-0.2cm}{\footnotesize +}
\end{center}

\end{minipage}};
\draw (50,21) node [anchor=north west][inner sep=0.75pt]   [align=left] {$t^{1}_{4}$ };
\draw (8,190.07) node [anchor=north west][inner sep=0.75pt]   [align=left] {2.- Compute \\ITE values};
\draw (110,250) node [anchor=north west][inner sep=0.75pt]   [align=left] {$ite^{1}_{4restaurant}$= 2};
\draw (167,213) node [anchor=north west][inner sep=0.75pt]   [align=left] {{\small From polarities to numbers* (Eq. 4)} };
\draw (230,250) node [anchor=north west][inner sep=0.75pt]   [align=left] {$ite^{1}_{4ambience}$ = 2};
\draw (350,250) node [anchor=north west][inner sep=0.75pt]   [align=left] {$ite^{1}_{4food}$ = 2};
\draw (10,267) node [anchor=north west][inner sep=0.75pt]   [align=left] {{\small*$\in$ [$-\tau, \tau$]=$[-2,2]$}};

\end{tikzpicture}

}
\caption{Example of \nuevo{generating the ITE values from the natural language review $t^{1}_{4}$.}}
\label{example_ite}
\end{figure}

\cz{Figure \ref{example_ite} illustrates the \nuevo{generation of the $ITE$ matrices} through an example. Specifically, it outlines the process of extracting the expert knowledge from the review provided by the expert $e_{1}$ based on the restaurant $x_{4}$.} The \nuevo{DOC-ABSADeepL model extract the opinions getting numerical evaluations regarding the \textit{restaurant}, \textit{food} and \textit{ambience} criteria, which perfectly matches with the written text and with the actual annotation label shown in Table \ref{exampleAnnotation}.} Thus, it is appreciated that $ITE$ matrices provide much more information, of high quality, than the $INE$ matrices.

\nuevo{The $INE$ matrices, collecting the numerical ratings provided directly by the experts, are shown in Figure \ref{cs_ine}.}
Since TripAdvisor allows to provide numerical evaluations just to the criteria \textit{restaurant},  \textit{food} and  \textit{service}, the \prev{columns} referring to the criteria \textit{drinks}, \textit{atmosphere} and \textit{location} contain missing values. Let us consider the evaluations provided by the expert \prev{$e_{1}$} to the restaurant $x_{4}$, \czm{which are collected into the column $ine^{1}_{4j}, j=1,...,6$}. The numerical evaluation provided to the criteria \textit{restaurant} takes the value 1, which means \textit{Very Good} \czm{according to TripAdvisor}. The numerical evaluations provided to the remaining criteria are not available since the expert does not provide them. Therefore, \geni{according to the} numerical evaluations, this expert provides information on just one criterion. However, as shown in Figure \ref{example_e1}, the element \prev{$t^{1}_{4}$} provides information about the criteria \textit{restaurant}, \textit{food} and \textit{ambience}. Through this example, we show that analyzing natural language reviews provide much more knowledge than analyzing just numerical evaluations. \nuevo{Likewise, the $INE$ matrices show that experts can numerically evaluate one, two or three criteria of the allowed ones. For example, the expert $e_{6}$ evaluates all possible criteria, \textit{restaurant}, \textit{food} and \textit{service} respectively, for the first alternative while evaluates only the \textit{restaurant} criterion for the second alternative.}
\begin{figure}[h!]
\prev{\tiny\[
INE^{1} =
    \begin{bmatrix}
    1 & NA & NA & NA & NA & NA\\
    2 & NA & NA & NA & NA & NA\\
    1 & NA & NA & NA & NA & NA\\
    1 & NA & NA & NA & NA & NA
    \end{bmatrix} 
\hspace{0.5cm}
INE^{2} =
    \begin{bmatrix}
    1 & NA & NA & NA & NA & NA\\
    2 & NA & NA & NA & NA & NA\\
    2 & NA & NA & NA & NA & NA\\
    1 & NA & NA & NA & NA & NA
    \end{bmatrix} 
\]
\[
INE^{3} =
    \begin{bmatrix}
    1 & 1 & 2 & NA & NA & NA\\
    2 & 2 & 2 & NA & NA & NA\\
    -1 & -1 & 0 & NA & NA & NA\\
    1 & 1 & 2 & NA & NA & NA\\
    \end{bmatrix}
\hspace{0.6cm}
INE^{4} =
    \begin{bmatrix}
    1 & 2 & 1 & NA & NA & NA\\
    2 & 2 & 1 & NA & NA & NA\\
    2 & 2 & 2 & NA & NA & NA\\
    -1 & -1 & -2 & NA & NA & NA
    \end{bmatrix}
\]
\[
INE^{5} =
    \begin{bmatrix}
    1 & 1 & 1 & NA & NA & NA\\
    2 & 2 & 2 & NA & NA & NA \\
    2 & 2 & 2 & NA & NA & NA \\
    2 & 1 & 2 & NA & NA & NA 
    \end{bmatrix}
\hspace{1cm}
INE^{6} =
    \begin{bmatrix}
    1 & 1 & 1 & NA & NA & NA\\
    0 & NA & NA & NA & NA & NA\\
    1 & 1 & 2 & NA & NA & NA\\
    2 & 2 & 2 & NA & NA & NA
    \end{bmatrix}
\]
\normalsize
}
\caption{Individual Numerical Evaluations matrices obtained by \nuevo{SA-MpMcDM}. \label{cs_ine}}
\end{figure}







\subsection{Alternative choice decisions}\label{selecting_best_exp}

\geni{Once the expert knowledge is extracted and numerically represented, the last phase of \nuevo{SA-MpMcDM} selects the best alternative ranking the restaurants of the TripR-2020 dataset. For this purpose, three tasks are carried out: 
\begin{enumerate*}[label=(\arabic*)]
\item individual aggregation;
\item collective aggregation; and 
\item exploitation. 
\end{enumerate*}}

\paragraph{\czm{Individual aggregation}}
To obtain the final \prev{individual preferences} of each expert for each alternative, \cz{we aggregate the $ITE^{k}$ and the $INE^{k}$ matrices getting the $IP^{k}$ matrices. Figure \ref{cs_ip} presents the $IP$ matrices obtained applying the Equation \ref{get_ip_kij}. Analyzing as example }the expert $e_{1}$ based on the restaurant $x_{4}$, we verify that the final numerical evaluation for the criterion \textit{restaurant} is not as bad as the expert indicates through the numerical evaluations, neither as good as the expert indicates through the written text. Therefore, the final numerical evaluation  \geni{consolidates} all the information that the expert provides through written texts and even numerical evaluations.


\begin{figure}[h!]
\prev{
\tiny
\[
IP^{1} =
    \begin{bmatrix}
    1.5 & NA & NA & NA & NA & 2 \\
    2 & 2 & NA & NA & NA & NA \\
    1.5 & NA & 2 & NA & NA & 2 \\
    1.5 & 2 & NA & NA & 2 & NA 
    \end{bmatrix} 
\hspace{0.65cm}
IP^{2} =
    \begin{bmatrix}
    1.5 & 2 & 2 & NA & 2 & 2 \\
    2 & 2 & NA & NA & NA & NA \\
    2 & 2 & 2 & NA & 2 & NA \\
    1.5 & 2 & 2 & NA & NA & NA 
    \end{bmatrix} 
\]
\[
IP^{3} =
    \begin{bmatrix}
    1.5 & 1.25 & 2 & 2 & 2 & NA \\
    2 & 2 & 2 & NA & 2 & NA \\
    0.5 & 0.28 & 0 & 2 & NA & NA \\
    1.5 & 1.5 & 2 & NA & 2 & NA 
    \end{bmatrix}
\hspace{0.4cm}
IP^{4} =
    \begin{bmatrix}
    1.5 & 2 & 1.5 & 2 & 2 & 2 \\
    2 & 2 & 1 & NA & NA & NA \\
    2 & 2 & 2 & 2 & 2 & 2 \\
    -0.3 & 0.25 & -1 & NA & NA & NA 
    \end{bmatrix}
\]
\[
IP^{5} =
    \begin{bmatrix}
    1.5 & 1.5 & 1 & NA & 2 & 2 \\
    2 & 2 & 2 & NA & NA & 2 \\
    2 & 2 & 2 & NA & 2 & NA \\
    2 & 1.5 & 2 & NA & 2 & NA 
    \end{bmatrix}
\hspace{0.8cm}
IP^{6} =
    \begin{bmatrix}
    1.5 & 1.5 & 1.5 & NA & 2 & NA \\
    1 & 2 & -2 & NA & 2 & NA \\
    1.5 & 1.5 & 2 & NA & 2 & NA \\
    2 & 2 & 2 & NA & NA & NA 
    \end{bmatrix}
\]
\normalsize
}
\caption{Individual Preference matrices obtained by \nuevo{SA-MpMcDM}.\label{cs_ip}}
\end{figure}


\paragraph{Collective aggregation} 
\geni{We aggregate all the $IP^{k}$ matrices for obtaining the final evaluation for each restaurant according to each criteri\cris{on}.} Figure \ref{cs_cp} presents the $CP$ matrix, which is obtained applying the average operator to the $IP$ matrices.\footnote{Figure \ref{example_matrices} shows the format of the matrix.} \cz{According to the obtained $CP$ matrix, }the restaurants \textit{Oxo Tower} and \textit{The Wolsely} are evaluated based on all the criteria. \textit{J. Sheekey} restaurant is evaluated 
according to all the criteria \geni{except} \textit{drinks}. Finally, \textit{The Ivy} \czm{is evaluated }
according to all the criteria \geni{except} \textit{drinks} and \textit{location}. \geni{The final evaluation of a restaurant only depends on the \cris{criteria evaluated} by the experts.}
\begin{figure}[h!]
\prev{
\scriptsize
\[
CP =
    \begin{bmatrix}
    1.5 & 1.65 & 1.6 & 2 & 2 & 2 \\
    1.83 & 2 & 0.75 & NA & 2 & 2 \\
    1.58 & 1.56 & 1.67 & 2 & 2 & 2 \\
    1.37 & 1.54 & 1.4 & NA & 2 & NA 
    \end{bmatrix}
\]
\normalsize
}
\caption{Collective Preference matrix obtained by \nuevo{SA-MpMcDM}.\label{cs_cp}}
\end{figure}



\paragraph{Exploitation} 
We compute the criteria weights according to \nuevo{weighting criteria through the attention of experts (see Equation \ref{get_weigths})}. Numerical ratings according to each criterion are directly provided by the experts, while opinions regarding each criterion are extracted \nuevo{by the DOC-ABSADeepL model}. The number of evaluations regarding each criterion are $n\_evaluations(restaurant) = 83$,  $n\_evaluations(food) = 105$, $n\_evaluations(service) = 34$, $n\_evaluations(drinks) = 8$, $n\_evaluations(ambience) = 30$ and  $n\_evaluations(location) = 11$. Then, the criteria weights are given by $\omega_{restaurant} = 0,306$, $\omega_{food}=0.387$, $\omega_{service}=0.125$, $\omega_{drinks}=0.03$, $\omega_{ambience}=0.111$ and $\omega_{location}=0.041$.

The final numerical evaluation of each restaurant is computed aggregating the evaluation of each criterion according to the criteria weights. The final preference $fp$ vector collecting the final numerical evaluations is given by the Equation \ref{get_fp_i}, getting \small
$fp_{Oxo Tower}=1.66$; $fp_{J. Sheekey}=1.73$; $fp_{The Wolseley}=1.65$; $fp_{The Ivy}=1.41$.
\normalsize
These values show that \begin{enumerate*}[label=(\arabic*)]
\item all the restaurants have very good quality; 
\item the best restaurant, based on the evaluations of the six selected experts, is J. Sheekey; and
\item The Oxo Tower and The Wolsely restaurants, which were evaluated according to all the criteria, are practically of the same quality.
\end{enumerate*}

The final ranking
\czm{, as shown in Table \ref{final_ranking},} is $x_{2} > x_{1} > x_{3} > x_{4}$, \textit{i.e.}, J. Sheekey $>$ Oxo Tower Restaurant Bar and Brasserie $>$ The Wolseley $>$ The Ivy. 
\begin{table}[!h]
\centering
\small
\begin{tabularx}{\textwidth}{cccc@{\hspace{8em}}c}
\toprule
\textbf{}	Oxo Tower & J. Sheekey & Wolseley & Ivy & Final Ranking\\
\midrule
 1.66 & 1.73 & 1.65 & 1.41 & $x_{2} > x_{1} > x_{3} > x_{4}$\\
\bottomrule
\end{tabularx}
\caption{Final ranking obtained by the \nuevo{SA-MpMcDM methodology} to solve the case of study \cz{collected into the TripR-2020 dataset}. Ratings belong to the interval $[-\tau, \tau] = [-2,2]$.}
\label{final_ranking}
\end{table}





\section{Behaviour analysis of the SA-MpMcDM methodology}\label{com_and_dis}

\nuevo{This section solves the case study exposed in Section \ref{case_study} thought \eugenio{three evaluation scenarios} to analyze the \nuevo{SA-MpMcDM methodology}. Section \ref{expected} exposes the solution based on the annotation of the evaluations from the TripR-2020 dataset to provide the ranking of the restaurants that best match the evaluations provided by the experts. Section \ref{just_numbers} provides the solution based on just numerical ratings to check the importance of adding natural language assessments. Section \ref{just_text} provides the solution based on just written texts to prove that numerical ratings improve the performance of SA-McMpDM.

\cz{All the evaluation scenarios analyze the TripR-2020 dataset\cris{,} which }
is composed by the restaurants $X = \{$\textit{Oxo Tower Restaurant, Bar and Brasserie}, \textit{J. Sheekey}, \textit{The Wolseley}, \textit{The Ivy}$\}$ and the experts $E=\{e_{1}, e_{2}, e_{3}, e_{4}, e_{5}, e_{6}\}$. The set of criteria is $C=\{$\textit{Restaurant}, \textit{Food}, \textit{Service}, \textit{Drinks}, \textit{Ambience}, \textit{Location}$\}$.} \cz{Each scenario solves the \nuevo{MpMcDM} problem providing the final ranking. Table \ref{ranking_all_v4} shows the three restaurant rankings corresponding to the previous three evaluation scenarios and the one returned by the \nuevo{SA-MpMcDM methodology}. We subsequently describe the three evaluation scenarios.
} 



\subsection{Solution based on the annotation of the evaluations}\label{expected}
\nuevo{The purpose of this section is to provide the final preference vector that best fit to the evaluations that the experts provide. Therefore,} this scenario provides the upper-bound result analyzing both numerical evaluations and written text. The case of study is solved simulating that the proposed \nuevo{SA-MpMcDM methodology takes into account the original opinions of the experts instead of the opinions obtained by the DOC-ABSADeepL model.}

The $ITE$ matrices are obtained \nuevo{extracting the opinions directly from} the annotated labels of TripR-2020 and computing the $ite$ values applying the Equation \ref{get_ite_kij}. \nuevo{For example, the actual annotation label of the sentence  \textquotedblleft Always good atmosphere and fun people watching.'' from the written text $t^{1}_{4}$ is given by the aspect \textquotedblleft atmosphere'', the category \textquotedblleft ambience'' and the polarity \textquotedblleft positive'' as shown in Table \ref{exampleAnnotation}.} The $INE$ matrices are the same than \czm{shown in Figure \ref{cs_ine}}, since the numerical evaluations provided by the experts are the same. Figure \ref{exp_ip} shows the $IP$ matrices\nuevo{, which are obtained applying equation \ref{get_ip_kij}.}

\begin{figure}[h!]
\prev{
\tiny
\[
IP^{1} =
    \begin{bmatrix}
    1.5 & NA & NA & NA & NA & 2 \\
    2 & 2 & 2 & NA & NA & NA \\
    1 & NA & 2 & NA & 2 & 2 \\
    1.5 & 2 & NA & NA & 2 & NA 
    \end{bmatrix}
\hspace{0.9cm}
IP^{2} =
    \begin{bmatrix}
    1.5 & 2 & 1 & NA & NA & 2 \\
    2 & 2 & NA & NA & NA & NA \\
    2 & NA & 2 & NA & 2 & NA \\
    0.5 & -2 & -2 & NA & NA & NA 
    \end{bmatrix} 
\]
\[
IP^{3} =
    \begin{bmatrix}
    1.5 & 0.93 & 0 & 2 & NA & NA \\
    1 & 1.86 & 2 & NA & 2 & NA \\
    0.5 & -1.1 & -0.5 & -2 & NA & NA \\
    1.5 & 0.7 & 2 & NA & 2 & NA 
    \end{bmatrix}
\hspace{0.45cm}
IP^{4} =
    \begin{bmatrix}
    1.5 & 2 & 1.5 & 2 & 2 & 2 \\
    2 & 2 & 1 & NA & NA & NA \\
    2 & 2 & 2 & NA & 2 & 2 \\
    -1.5 & -1.5 & -2 & NA & -2 & NA 
    \end{bmatrix}
\]
\[
IP^{5} =
    \begin{bmatrix}
    1.5 & 1.5 & 1 & NA & NA & 2 \\
    1.5 & 2 & 2 & NA & NA & 0 \\
    2 & 2 & 2 & NA & 0.67 & NA \\
    2 & 1.5 & 2 & NA & 2 & NA 
    \end{bmatrix} 
\hspace{1.6cm}
IP^{6} =
    \begin{bmatrix}
    1.5 & 1.5 & 1.5 & NA & 2 & NA \\
    0 & 2 & 2 & NA & 1 & NA \\
    1.3 & 1.5 & 2 & NA & 2 & 2 \\
    2 & 2 & 2 & NA & 2 & NA 
    \end{bmatrix}
\]
\normalsize
}
\caption{Individual Preference matrices obtained \nuevo{considering the annotation of the evaluations.}\label{exp_ip}}
\end{figure}

\cz{Figure \ref{exp_cp} shows the $CP$ matrix, which is }
obtained applying the average operator to the $IP$ matrices. The criteria weights are obtained by means of the Equation \ref{get_weigths}, considering the original opinions getting $\omega_{restaurant} = 0,339$, $\omega_{food}=0.322$, $\omega_{service}=0.159$, $\omega_{drinks}=0.021$, $\omega_{ambience}=0.113$ and $\omega_{location}=0.046$. Applying the Equation \ref{get_fp_i}, the $fp$ vector is obtained getting $fp_{Oxo Tower}=1.538$, $fp_{J. Sheekey}=1.572$, $fp_{The Wolseley}=1.349$ and $fp_{The Ivy}=0.683$. 
\begin{figure}[!h]
\prev{
\tiny
\[
CP =
    \begin{bmatrix}
    1.5 & 1.59 & 1 & 2 & 2 & 2 \\
    1.42 & 1.98 & 1.8 & NA & 1.5 & 0 \\
    1.47 & 1.1 & 1.58 & -2 & 1.73 & 2 \\
    1 & 0.45 & 0.4 & NA & 1.2 & NA 
    \end{bmatrix}
\]
\normalsize
}
\caption{Collective Preference matrix obtained \nuevo{considering the annotation of the evaluations.}\label{exp_cp}}
\end{figure}

\nuevo{Therefore, the final ranking of the restaurants for the solution based on the annotation of the evaluations is $x_{2} > x_{1} > x_{3} > x_{4}$, that is, J. Sheekey $>$ Oxo Tower Restaurant Bar and Brasserie $>$ The Wolseley $>$ The Ivy.}

\subsection{Solution based on just numerical ratings}\label{just_numbers}
\nuevo{The aim of this section is to simulate the solution obtained by a traditional MpMcDM model that analyzes just the numerical evaluations. Therefore, this scenario solves the case of study by means of a simplification of the SA-MpMcDM methodology that only analyze the numerical ratings.} Since TripAdvisor does not allow experts to provide numerical evaluations to the criteria \textit{drinks}, \textit{ambience} and \textit{location}, we \new{reduce the set of criteria to} $C=\{$\textit{Restaurant}, \textit{Food}, \textit{Service}$\}$. 

The $ITE$ matrices do not exist in this model\geni{, because  the textual evaluations are not analyzed}. The $INE$ matrices are the same than \czm{shown in Figure \ref{cs_ine}}, since the numerical evaluations provided by the experts are the same. Then, the $IP$ matrices are the $INE$ matrices. 

\prev{
\cz{Figure \ref{just_num_cp} shows the $CP$ matrix, which is obtained applying the average operator to the $INE$ matrices.} The criteria weights are obtained by means of a simplification of the Equation \ref{get_weigths} counting just the number of numerical ratings, getting $\omega_{restaurant} = 0.444$, $\omega_{food}=0.278$ and $\omega_{sevice}=0.278$.} Applying the Equation \ref{get_fp_i}, the $fp$ vector is obtained getting $fp_{Oxo Tower}=1.139$, $fp_{J. Sheekey}=1.759$, $fp_{The Wolseley}=1.213$ and $fp_{The Ivy}=0.931$. 

\begin{figure}[h!]
\prev{
\tiny
\[
CP =
    \begin{bmatrix}
    1 & 1.25 & 1.25 & NA & NA & NA \\
    1.67 & 2 & 1.67 & NA & NA & NA \\
    1.17 & 1 & 1.5 & NA & NA & NA \\
    1 & 0.75 & 1 & NA & NA & NA 
    \end{bmatrix}
\]
\normalsize
}
\caption{Collective Preference matrix obtained analyzing just numerical evaluations.\label{just_num_cp}}
\end{figure}

\nuevo{Then, the final ranking of the restaurants for the solution based on just numerical ratings is $x_{2} > x_{3} > x_{1} > x_{4}$, that is, J. Sheekey $>$ The Wolseley $>$ Oxo Tower Restaurant Bar and Brasserie $>$ The Ivy.}

\subsection{Solution based on just written texts}\label{just_text}
\nuevo{This Section aims at proving the importance of allowing experts to provide numerical ratings to evaluate even though they can provide natural language evaluations. Hence, this scenario solves the case of study by means of a simplification of the SA-MpMcDM methodology that only analyze the written texts.}

The $ITE$ matrices are the same than \czm{shown in Figure \ref{cs_ite}}, since the written texts provided by the experts are the same. The $INE$ matrices do not exist in this model, since numerical ratings are not analyze\cris{d}. Then, the $IP$ matrices are the $ITE$ matrices. 

\prev{
\cz{Figure \ref{just_text_cp} shows the $CP$ matrix, which is obtained applying the average operator to the $ITE$ matrices.} The criteria weights are obtained by means of a simplification of Equation \ref{get_weigths} counting just the number of opinions \nuevo{obtained by the DOC-ABSADeepL model} getting $\omega_{restaurant} = 0,272$, $\omega_{food}=0.415$, $\omega_{service}=0.088$, $\omega_{drinks}=0.04$, $\omega_{ambience}=0.138$ and $\omega_{location}=0.051$. Applying the Equation \ref{get_fp_i}, the $fp$ vector is obtained getting $fp_{Oxo Tower}=1.958$, $fp_{J. Sheekey}=1.81$, $fp_{The Wolseley}=1.954$ and $fp_{The Ivy}=1.667$.} 

\begin{figure}[h!]
\prev{
\tiny
\[
CP =
    \begin{bmatrix}
    2 & 1.9 & 2 & 2 & 2 & 2 \\
    2 & 2 & 0.67 & NA & 2 & 2 \\
    2 & 1.89 & 2 & 2 & 2 & 2 \\
    1.73 & 1.9 & 1.5 & NA & 2 & NA 
    \end{bmatrix}
\]
\normalsize
}
\caption{Collective Preference matrix obtained analyzing just written texts. \label{just_text_cp}}
\end{figure}

\nuevo{Therefore, the final ranking of the restaurants for the solution based on just written texts is $x_{1} > x_{3} > x_{2} > x_{4}$, that is, Oxo Tower Restaurant Bar and Brasserie $>$  The Wolseley $>$  J. Sheekey $>$ The Ivy.}




\subsection{Global Analysis and lessons learned}

Table \ref{ranking_all_v4} shows the results reached by the previous evaluation scenarios and by the \nuevo{SA-MpMcDM methodology}. The main conclusion is that it is necessary to deal both written texts and the numerical evaluations to have the highest possible quality in the result. In more detail, \eugenio{we conclude that}:  

\begin{table}[!h]
\centering
\small
\begin{tabularx}{\textwidth}{@{}X@{\hspace{1em}}ccc@{\hspace{1em}}cc@{}}
\toprule
\textbf{}	& \footnotesize{Oxo Tower} & \footnotesize{J. Sheekey} & \footnotesize{Wolseley} & \footnotesize{Ivy} & \footnotesize{Final Ranking}\\
\midrule
\scriptsize{Annotated evaluations}	& 1.54 & 1.57 & 1.35 & 0.68	& \footnotesize{$x_{2} > x_{1} > x_{3} > x_{4}$}\\
\scriptsize{Only numerical eval.} &  1.14 & 1.76 & 1.21 & 0.93 & \footnotesize{$x_{2} > x_{3} > x_{1} > x_{4}$}\\
\scriptsize{Only text eval.}	&  1.96 & 1.81 & 1.95 & 1.67 & \footnotesize{$x_{1} > x_{3} > x_{2} > x_{4}$}\\
\scriptsize{num.+text \tiny{(SA-MpMcDM)}}
&  1.66 & 1.73 & 1.65 & 1.41 & \footnotesize{$x_{2} > x_{1} > x_{3} > x_{4}$}\\
\bottomrule
\end{tabularx}
\caption{\nuevo{Results obtained by the different scenarios from the SA-MpMcDM methodology} to solve the case of study collected into the TripR-2020 dataset. Ratings belong to the interval $[-\tau, \tau] = [-2,2]$.}
\label{ranking_all_v4}
\end{table}

    \begin{enumerate}[noitemsep]
        \item \prev{\textbf{The final evaluations obtained by \nuevo{SA-MpMcDM} agree with the \nuevo{solution based on the actual annotation labels}.} The final ranking obtained by \nuevo{SA-MpMcDM}, which aggregate numerical and written text evaluations, matches with the expected result. By contrast, models that handle only numerical or textual evaluations do not match the expected ranking. }
        \item \textbf{The final evaluations obtained by \nuevo{SA-MpMcDM} agree with what experts would expect when reading the reviews.} 
        Reading \czm{the original reviews from the TripR-2020 dataset }we conclude that \begin{enumerate*}[label=(\arabic*)]
            \item \textit{The J. Sheekey} is the best restaurant since it only has a negative opinion; 
            \item \textit{The Ivy} has the worst rating, and therefore is the worst restaurant, since one expert provides a very negative review; and 
            \item \textit{The Oxo Tower} and \textit{The Wolsely} restaurants have very similar and good opinions, however, many details about \textit{The Oxo Tower} are analyzed in a positive way, which makes it to be in the lead.
        \end{enumerate*}
        \item \textbf{\nuevo{SA-MpMcDM} allows analyzing any criterion, while \prev{traditional models allow} analyzing just pre-defined criteria.} The proposed methodology allows experts to evaluate any aspect related to any criterion. An expert may express her opinion on another criterion than the six established criteria. Simply, the \nuevo{DOC-ABSADeepL model} will have to be trained with this extra criterion. Therefore, we do not limit the expert to evaluate on \nuevo{pre-defined} criteria. In contrast, traditional models  
        limit the criteria that can be evaluated by experts. In the case study, using the traditional model, experts can only provide numerical evaluations to three of the six criteria that can be evaluated using the \nuevo{SA-MpMcDM} methodology.
        \item \textbf{\nuevo{SA-MpMcDM} provides so much more information that \prev{rest of models}.} Thanks to the proposed methodology experts feel free to express opinions regarding any criteri\cris{on} of a restaurant. It is much easier and more comfortable for experts to express their opinions on writing text than providing numerical values. Experts are encouraged to evaluate more criteria when they can freely express their opinion\cris{s} without being limited to numerical values. This can be verified by analyzing, for example, expert $e_{2}$ evaluations. As shown in Figure \prev{\ref{cs_ine},} this expert only provides numerical evaluations to the overall \textit{restaurant} criteri\cris{on}. 
        However, as shown in Figure \prev{\ref{cs_ite},} this expert provides opinions in the natural language written texts about \textit{food} and \textit{service} criteria for almost all the restaurants.
        \item \textbf{\prev{The worst restaurant is the same for all scenarios.}} The Ivy restaurant has been selected as the worst, although it is also high quality, by means of all scenarios. \nuevo{This means that, in general terms, what an expert express through written texts agrees with what she expresses through numerical ratings.}
        \item \prev{\textbf{Experts are more critical when they only provide numerical evaluations.} Results obtained by \nuevo{scenarios} dealing with written text are higher than \nuevo{results from scenarios} dealing just with numerical evaluations. Therefore, it follows that experts strongly penalize negative situations if they are not allowed to express themselves freely. }
    \end{enumerate}

\section{Conclusions and future work}\label{conclusions}

This paper presents the \nuevo{SA-MpMcDM} methodology which incorporates SA tasks \segrev{together with deep learning} to solve \nuevo{MpMcDM} problems \segrev{for smarter decision aid. This is a data driven decision making approach guided by artificial intelligence approaches for learning and transform data (linguistic reviews) into decisions, as a smarter decision aid approach.} 

In contrast to traditional models, \nuevo{SA-MpMcDM} allows experts to evaluate without imposing them any restriction related to how to express their evaluations. Specifically, experts can evaluate the alternatives in natural language with \nuevo{SA-MpMcDM}, and they can also perform numerical evaluations. SA methods allow to extract the expert knowledge from the natural language evaluations. \nuevo{In particular, the SA-MpMcDM methodology incorporates an end-to-end multi-task deep learning model, named DOC-ABSADeepL, to extract the opinions from the texts written in natural language.}


We evaluated the \nuevo{SA-MpMcDM methodology} on a real \nuevo{MpMcDM} \segrev{case study of} selecting the best restaurant according to the reviews posted on TripAdvisor. We compiled the TripR-2020 dataset for the evaluation, and we release it for evaluating other prospective DM models based on the SA-MpMcDM methodology.


The results allow to conclude that:
\begin{enumerate}[noitemsep]
    \item The use of the natural language evaluations enhances the performance of DM models.
    \item The \nuevo{SA-MpMcDM methodology} allows experts to evaluate any criterion. 
    \item The \nuevo{SA-MpMcDM methodology} provides more information than traditional DM models. 
    \item The results obtained by \nuevo{SA-MpMcDM} agree with expected results on the real \nuevo{MpMcDM} problem. 
    \item The evaluations of experts are usually more negative when they use numerical evaluations than when they express them in natural language.
\end{enumerate}

Therefore, we conclude that the combination of natural language and numerical information by means of the \nuevo{SA-MpMcDM methodology} is essential to reach higher quality decisions in \nuevo{MpMcDM} models.

\new{We will work on making the SA-MpMcDM methodology support large scale DM problems using sparse representation \cite{runxi2019}, and we will evaluate it with all the reviews of the TripR-2020 dataset, which were written by 1,428 experts. The main challenge will be the processing of those alternatives with sparse evaluations, because not all the alternatives are evaluated by all the experts. Moreover, we will study the consideration of the correlation among criteria, since they tend to be dependent and redundant in complex models \cite{liaochoquet, zhu2016multi}. We will also plan to extend the methodology taking into account the time in which the experts carry out the evaluations, so that the older evaluations are less important.}




\prev{
\section*{Acknowledgements}
This work was partially supported by the Spanish Ministry of Science and Technology under the projects TIN2017-89517-P and \nuevo{PID2019-103880RB-I00}. Cristina Zuheros and Eugenio Martínez Cámara were supported by the Spanish Government by the FPI Programme (PRE2018-083884) and Juan de la Cierva Incorporación (IJC2018-036092-I) respectively.}

\bibliographystyle{plain}
\bibliography{SA-MpMcDM.bib}
\new{
\newpage
\appendix
\section{Abbreviations and notation}
\label{appendixA}
}
\begin{table}[!h]
\centering
\small
\begin{tabularx}{\textwidth}{@{}lX@{}}
\toprule
MpMcDM		& Multi-person Multi-criteria Decision Making\\
SA		& Sentiment Analysis\\
SA-MpMcDM		& Sentiment Analysis based Multi-person Multi-criteria Decision Making methodology\\
ABSA & Aspect-Based Sentiment Analysis\\
DOC-ABSADeepL & Model for distilling opinions and criteria using an ABSA based deep learning model\\
$e_{k}, k=1,...,l$ & Expert $k$\\
$x_{i}, i=1,...,n$ & Alternative $i$\\
$c_{j}, j=1,...,m$ & Criteria $j$\\
$\omega_{j}, j=1,...,m$ & Criteria weight $j$\\
$T^{k}$ & Vector collecting the written text evaluations of $e_{k}$\\
$t^{k}_{i}$ & Written text evaluations of $e_{k}$ to $x_{i}$\\
$\tau$ & Numerical ratings belong to the interval $[-\tau, \tau]$\\
$\{w_{1},...,w_{s}\}$ & Input sentence ($s$ words) to the DOC-ABSADeepL model\\
$INE$ & Individual Numerical Evaluation matrix\\ 
$ine^{k}_{ij}$ & Individual numerical evaluation from $e_{k}$ to $x_{i}$ according to $c_{j}$\\  
$ITE$ & Individual Textual Evaluation matrix\\
$ite^{k}_{ij}$ & Individual textual evaluation from $e_{k}$ to $x_{i}$ according to $c_{j}$\\
$IP$ & Individual Preference matrix  aggregating $INE$ and $ITE$\\ 
$ip^{k}_{ij}$ & Individual preference from $e_{k}$ to $x_{i}$ according to $c_{j}$\\  
$CP $ & Collective Preference matrix aggregating $IP$ matrices \\
$cp_{ij}$ & Collective preference from all experts to $x_{i}$ according to $c_{j}$\\  
$FP$ & Final Preference vector \\
$fp_{i}$ & Final preference from all experts to $x_{i}$ according to all criteria\\  
\bottomrule
\end{tabularx}
\end{table}

\end{document}